\PassOptionsToPackage{pdfa}{hyperref}
\PassOptionsToPackage{inline}{enumitem}
\documentclass[acmsmall]{acmart}

\usepackage{acmcs-skeskislr}

\usepackage{acmcs-prologue}

\begin{abstract}
In this paper we focus on the opacity issue of sub-symbolic machine learning predictors by promoting two complementary activities---namely, \emph{symbolic knowledge extraction} (SKE) and \emph{injection} (SKI) from and into sub-symbolic predictors.
We consider as symbolic any language being intelligible and interpretable for both humans and computers. 
Accordingly, we propose general meta-models for both SKE and SKI, along with two taxonomies for the classification of SKE and SKI methods. 
By adopting an explainable artificial intelligence (XAI) perspective, we highlight how such methods can be exploited to mitigate the aforementioned opacity issue. 
Our taxonomies are attained by surveying and classifying existing methods from the literature, following a systematic approach, and by generalising the results of previous surveys targeting specific sub-topics of either SKE or SKI alone. 
More precisely, we analyse \skecount{} methods for SKE and \skicount{} methods for SKI, and we categorise them according to their purpose, operation, expected input/output data and predictor types. 
For each method, we also indicate the presence/lack of runnable software implementations.
Our work may be of interest for data scientists aiming at selecting the most adequate SKE/SKI method for their needs, and also work as suggestions for researchers interested in filling the gaps of the current state of the art, as well as for developers willing to implement SKE/SKI-based technologies.
\end{abstract}



\begin{document}

\maketitle

\section{Introduction}
In the context of artificial intelligence (AI), more and more critical applications that rely on machine learning (ML) are being developed.
This promotes a data-driven approach to the engineering of intelligent computational systems where hard-to-code tasks are (semi-)automatically learned from data rather than manually programmed by human developers.
Tasks that can be learned this way range from text \citep{otter_survey_2021} to speech \citep{nassif_speech_2019} or image recognition \citep{zhao_object_2019}, stepping through time series forecasting, clustering, and so on.
Applications are manifold, and make our life easier in many ways---e.g., via speech-to-text applications, email spam and malware filtering, customer profiling, automatic translation, virtual personal assistants, and so forth. 

Learning, in particular, is automated via ML algorithms, often implying \emph{numeric} processing of data---which in turn enables the detection of fuzzy patterns or statistically-relevant regularities in the data, that algorithms can learn to recognise.
This is fundamental to support the automatic acquisition of otherwise hard-to-formalise behaviours for computational systems.
However, flexibility comes at the cost of poorly-\emph{interpretable} solutions, as state-of-the-art \emph{sub-symbolic} predictors -- such as neural networks -- are often exploited behind the scenes.

These predictors are commonly characterised by opacity \cite{Burrell2016,Lipton2018}, as the interplay among the complexity of the data and the algorithms they are trained upon/with makes it hard for humans to understand their behaviour.
Hence, by `interpretable' we here mean that the expert human user may observe the computational system and understand its behaviour.
Even though the property is not always required, there exist safety-, value-, or ethic-critical applications where humans must be in full control of the computational systems supporting their decisions or aiding their actions.
In those cases, the lack of interpretability is a no-go.

State-of-the-art ML systems rely on a collection of well-established data mining predictors, such as neural networks, support vector machines, decision trees, random forests, or linear models.
Despite the latter sorts of predictors being often considered as interpretable in the general case, as the complexity of the problem at hand increases (e.g., dimensionality of the available data) trained predictors become more complex, hence harder to contemplate, and therefore less interpretable.
Nevertheless, these mechanisms have penetrated the modern practices of data scientists because of their flexibility, and expected effectiveness---in terms of predictive performance.
Unfortunately, a number of experts have empirically observed an inverse proportionality relation among interpretability and predictive performance~\citep{expectation-extraamas2021,Rudin2019}.
This is the reason why data-driven engineering efforts targeting critical application scenarios nowadays have to choose between predictive performance and interpretability as their priority: we call this the \emph{interpretability/performance trade-off}.

In this paper we focus on the problem of working around the interpretability/per\-for\-man\-ce trade-off.
We do so by promoting two complementary activities, namely \emph{symbolic knowledge extraction} (SKE) and \emph{injection} (SKI) from and into sub-symbolic predictors.
In both cases, `symbolic' refers to the way knowledge is represented.
In particular, we consider as symbolic any language that is intelligible and interpretable for both human beings and computers.
This includes a number of logic formalisms, and excludes the fixed-sized tensors of numbers commonly exploited in sub-symbolic ML.

Intuitively, SKE is the process of distilling the knowledge a sub-symbolic predictor has grasped from data into symbolic form.
This can be exploited to provide explanations for otherwise poorly-interpretable sub-symbolic predictors.
More generally, SKE enables the \emph{inspection} of the sub-symbolic predictors it is applied to, making it possible for the human designer to figure out how they behave.
Conversely, SKI is the inverse process of letting a sub-symbolic predictor follow the symbolic knowledge possibly encoded by its human designers.
It enables a higher degree of \emph{control} over a sub-symbolic predictor and its behaviour, by constraining it with human-like common-sense---suitably encoded into symbolic form.

Apart from insights, notions such as SKE and SKI have rarely been described in general terms into the scientific literature---despite the multitude of methods falling under their umbrellas.
Hence, the aim of this paper is to provide general definitions and descriptions of these topics, other than providing durable taxonomies for categorising present and future SKE/SKI methods.
Arguably, these contributions should take into account the widest possible portion of scientific literature, so as to avoid subjectivity.
Accordingly, in this paper we propose a systematic literature review (SLR) following the three-folded purpose of
\begin{inlinelist}
    \item collecting and categorising existing methods for SKE and SKI into clear taxonomies,
    \item providing a wide overview of the state of the art and technology, 
    and
    \item detecting open research challenges and opportunities.
\end{inlinelist}
In particular, we analyse \skecount{} methods for SKE and \skicount{} methods for SKI, classifying them according to their purpose, operation, expected input/output data and predictor types.
For each method, we also probe the existence/lack of software implementations.

To the best of our knowledge, our survey is the only \emph{systematic} work focusing on \emph{both} SKE and SKI algorithms.
Furthermore, w.r.t.\ other surveys on these topics, our SLR collects the greatest number of methods.
In doing so, we elicit a meta-model for SKE (resp.\ SKI) according to which existing and future extraction (resp.\ injection) methods can be categorised and described.
Our taxonomies may be of interest for data scientists willing to select the most adequate SKE/SKI method for their needs, and also work as suggestions for researchers interested in filling the gaps of the current state of the art, or developers willing to implement SKE or SKI software technologies.

Accordingly, the remainder of this paper is organised as follows.
\Cref{sec:background} recalls the state of the art for machine learning, symbolic AI, and explainable AI (XAI), aimed at providing readers with a fast-track access to most of the concepts and terms used in the paper.
\Cref{sec:preliminaries} delves into the details of what we mean by SKE and SKI, and explains how this SLR is conducted: there, we declare our research questions and describe our research methodology.
Then, \Cref{sec:results} answers our research questions, summarising the results of the analysis of the surveyed literature.
The same results are then discussed in \Cref{sec:discussion}, where major challenges and opportunities are elicited.
Finally, \Cref{sec:conclusion} concludes the paper.

\section{Background}
\label{sec:background}

\subsection{Machine Learning}
A widely adopted definition of machine learning by \cite{Mitchell1997} states:
\begin{displayquote}\itshape\itshape
    a computer program is said to learn from experience $E$ with respect to some class of tasks $T$ and performance measure $P$ if its performance at tasks in $T$, as measured by $P$, improves with experience $E$.
\end{displayquote}
This definition is very loose, as it does not specify
\begin{inlinelist}
    \item what are the possible tasks,
    \item how performance is measured in practice,
    \item how/when experience should be provided to tasks,
    \item how exactly the program is supposed learn, and
    \item under which form learnt information is represented.
\end{inlinelist}
Accordingly, depending on the particular ways these aspects are tackled, a categorisation of the approaches and techniques enabling software agents to learn may be drawn.

Three major approaches to ML exist: namely, \emph{supervised}, \emph{unsupervised}, and \emph{reinforcement} learning.
Each approach is tailored on a well-defined pool of tasks, which may, in turn, be applied in a wide range of use case scenarios.
Accordingly, differences among three major approaches can be understood by looking at the sorts of tasks $T$ they support -- commonly consisting of the estimation of some unknown relation --, and how experience $E$ is provided to the learning algorithm.

In supervised learning, the learning task consists of finding a way to approximate an unknown relation given a sampling of its items---which constitute the experience.
In unsupervised learning, the learning task consists of finding the best relation for a sample of items -- which constitute the experience --, following a given optimality criterion intensionally describing the target relation.
In reinforcement learning, the learning task consists of letting an agent estimate optimal plans given the reward it receives whenever it reaches particular goals.
There, the rewards constitutes the experience, while plans can be described as relations among the possible states of the world, the actions to be performed in those states, and the rewards the agents expects to receive from those actions.

Several practical AI problems -- such as image recognition, financial and medical decision support systems -- can be reduced to \emph{supervised} ML---which can be further grouped in terms of either \emph{classification} or \emph{regression} problems \citep{twala2010,smlreview-faia160}.
Within the scope of sub-symbolic supervised ML, a \emph{learning algorithm} is commonly exploited to approximate the specific nature and shape of an unknown \emph{prediction} function (or \emph{predictor}) $\pi^*: \mathcal{X} \rightarrow \mathcal{Y}$, mapping data from an input space $\mathcal{X}$ into an output space $\mathcal{Y}$.
There, common choices for both $\mathcal{X}$ and $\mathcal{Y}$ are, for instance, the set of vectors, matrices, or tensors of numbers of a given size---hence the sub-symbolic nature of the approach.

%
Without loss of generality, in the following we refer to items in $\mathcal{X}$ as $n$-dimensional vectors denoted as $\mathbf{x}$, whereas items in $\mathcal{Y}$ are $m$-dimensional vectors denoted as $\mathbf{y}$---despite matrices or tensors may be suitable choices as well.

To approximate function $\pi^*$, supervised learning assumes that a \emph{learning algorithm} is in place.
This algorithm computes the approximation by taking into account a number $N$ of \emph{examples} of the form $(\mathbf{x}_i,\mathbf{y}_i)$ such that $\mathbf{x}_i \in X \subset \mathcal{X}$, $\mathbf{y}_i \in Y \subset \mathcal{Y}$, and $|X| \equiv |Y| \equiv N$.
There, the set $D = \{ (\mathbf{x}_i,\mathbf{y}_i) \mid \mathbf{x}_i \in X, \mathbf{y}_i \in Y \}$ is called \emph{training} set, and it consists of $(n+m)$-dimensional vectors.
The dataset can be considered as the concatenation of two matrices, namely the $N \times n$ matrix of \emph{input} data ($X$) and the $N \times m$ matrix of \emph{expected output} data ($Y$).
There, each $\mathbf{x}_i$ represents an instance of the input data for which the expected output value $\mathbf{y}_i \equiv \pi^*(\mathbf{x}_i)$ is known or has already been estimated.
Notably, such sorts of ML problems are said to be `supervised' \emph{because} the expected outputs $Y$ are available.
Furthermore, the function approximation task is called
%
    \textbf{regression} if the components of $Y$ consist of continuous or numerable -- i.e., \emph{infinite} -- values,
    \textbf{classification} if they consist of categorical -- i.e., \emph{finite} -- values.

\subsubsection{On the Nature of Sub-Symbolic Data}

ML methods, and sub-symbolic approaches in general, represent data as (possibly multi-dimensional) \emph{arrays} (e.g., vectors, matrices, or tensors) of real numbers, and knowledge as functions over data.
This is particularly relevant as opposed to \emph{symbolic} knowledge representation approaches, which represent data via logic formul\ae{} (cf.\ \Cref{subsec:sym-ai}).

In spite of the fact that numbers are technically symbols as well, we cannot consider arrays and their functions as means for symbolic knowledge representation (KR).
Indeed, according to \cite{Gelder90}, to be considered as symbolic, KR approaches should
\begin{enumerate*}[label=\emph{(\alph{*})}]
    \item involve a set of symbols,
    \item\label{item:combination} which can be combined (e.g., concatenated) in possibly infinite ways, following precise grammatical rules, and
    \item\label{item:meaning} where both elementary symbols and any admissible combination of them can be assigned with \emph{meaning}---i.e., each symbol can be mapped into some entity from the domain at hand.
\end{enumerate*}
Below, we discuss how sub-symbolic approaches most typically do not satisfy requirements \ref{item:combination} and \ref{item:meaning}.

\paragraph{Vectors, matrices, tensors}

Multi-dimensional arrays are the basic brick of sub-symbolic data representation.
More formally, a $D$-order array consists of an ordered container of real numbers, where $D$ denotes the amount of indices required to locate each single item into the array.
%
%
We may refer to 1-order arrays as \emph{vectors}, 2-order arrays as \emph{matrices}, and higher-order arrays as \emph{tensors}.
%

In any given sub-symbolic data-representation task leveraging upon arrays, information may be carried by both
\begin{inlinelist}
    \item the actual numbers contained into the array, and
    \item their location into the array itself.
\end{inlinelist}
In practice, the actual dimensions $(d_1 \times \ldots \times d_D)$ of the array play a central role as well.
Indeed, sub-symbolic data processing is commonly tailored on arrays of \emph{fixed} sizes---meaning that the actual values of $d_1, \ldots, d_D$ are chosen at design time and never changed after that.
This violates requirement \ref{item:combination} above, hence we define sub-symbolic KR as the task of expressing data in the form of \emph{rigid} arrays of \emph{numbers}.

\paragraph{Local vs.\ distributed}

When data is represented in the form of numeric arrays, the whole representation may be \emph{local} or \emph{distributed} \citep{Gelder90}.
In local representations, each single number into the array is characterised by a well-delimited meaning---i.e., it is measuring or describing a clearly-identifiable concept from a given domain.
Conversely, in distributed representations, each single item of the array is nearly meaningless, unless it is considered along with its neighbourhood---i.e., any other item which is `close' in the indexing space of the array, according to some given notion of closeness.
So, while in local representations the location of each number in the array is mostly negligible, in distributed representations it is of paramount importance. 
Notably, distributed representations violate the aforementioned requirement \ref{item:meaning}.
In recent literature, authors call `sub-symbolic' those predictors who rely on distributed representations of data.

\subsubsection{Overview on ML Predictors}

Depending on the predictor family of choice, the nature of the admissible hypothesis spaces and learning algorithms may vary dramatically, as well as the predictive performance of the target predictor, and the whole efficiency of learning.

In the literature of machine learning, statistical learning, and data mining, a plethora of learning algorithms have been proposed along the years.
Because of the `no free lunch' (NFL) theorem \citep{DolpertM97}, however, no algorithm is guaranteed to outperform the others in all possible scenarios.
For this reason, the literature and the practice of data science keeps leveraging on algorithms and methods whose first proposal was published decades ago.
Most notable algorithms include, among the many others, (deep) neural networks (NN), decision trees (DT), (generalised) linear models, nearest neighbours, support vector machines (SVM), and random forests.

These algorithms can be categorised in several ways, for instance depending
\begin{inlinelist}
    \item on the supervised learning task they support (classification vs.\ regression), or
    \item on the underlying strategy adopted for learning (e.g., gradient descent, least square optimisation).
\end{inlinelist}

Some learning algorithms (e.g., NN) naturally target regression problems -- despite being adaptable to classification, too --, whereas others (e.g., SVM) target classification problems---while being adaptable to regression as well.
Similarly, some target multi-dimensional outputs ($\mathbf{y} \in \mathbb{R}^m$, and $m>1$), whereas others target mono-dimensional outputs ($m = 1$).
Regressors are considered as the most general case, as other learning tasks can usually be defined in terms of mono-dimensional regression.
%

The learning strategy is inherently bound to the predictor family of choice.
NN, for instance, are trained via back-propagation \citep{Rumelhart1986} and stochastic gradient descent (SGD), generalised linear models via Gauss' least squares method, decision trees via methods described in \citep{breiman1984classification}, etc.
Even though all the aforementioned algorithms may appear interchangeable in principle -- because of the NFL theorem --, their malleability is very different in practice.
For instance, the least square method involves inverting matrices of order $N$ -- where $N$ is the amount of available examples in the training set --, making the computational complexity of learning more than quadratic in time.
Furthermore, in practice, convergence of the method is not guaranteed in the general case; instead, it is guaranteed for generalised linear models---hence it is not adopted elsewhere.
Thus, learning by least square optimisation may become impractical for big datasets or for predictor families outside the scope of generalised linear models.
Conversely, the SGD method involves arbitrarily-sized subsets of the dataset (a.k.a.\ batches) to be processed a finite (i.e., controllable) amount of times.
Hence, the complexity of SGD can be finely controlled and adapted to the computational resources at hand---e.g., by making the learning process incremental, and by avoiding all data to be loaded in memory.
Moreover, SGD can be applied to several sorts of predictor families (there including NN and generalised linear models), as it only requires the target function to be differentiable w.r.t.\ its parameters.
For all these reasons, despite the lack of optimality guarantees, SGD is considered as very effective, scalable, and malleable in practice, hence it is extensively exploited in the modern data science applications.

In the remainder of this subsection we focus on two families of predictors -- namely, DT and NN --, and their respective learning methods.
We focus precisely on them because they are related to many surveyed SKE/SKI methods.
DT are noteworthy because of their user friendliness, whereas NN are mostly popular because of their predictive performance and flexibility.

\paragraph{Decision trees}
\label{par:decision-trees}


Decision trees are particular sorts of predictors supporting both classification and regression tasks.
In their learning phase, the input space is recursively \emph{partitioned} through a number of splits (a.k.a.\ \emph{decisions}) based on the input data $X$, in such a way that the prediction in each partition is constant, and the error w.r.t.\ the expected outputs $Y$ is minimal, while keeping the total amount of partitions low as well.
The whole procedure then synthesises a number of \emph{hierarchical} decision rules to be followed whenever the prediction corresponding to any $x \in \mathcal{X}$ must be computed.
In the inference phase, decision rules are orderly evaluated from the root to a leaf, to select the portion of the input space  $\mathcal{X}$ containing $x$.
As each leaf corresponds to a single portion of the input space, the whole procedure results in a single prediction for each $x$.

Unlike other families of predictors, the peculiarity of DT lies in the particular outcome of the learning process -- namely, the \emph{tree} of decision rules -- which is straightforwardly intelligible for humans and graphically representable in 2D charts.
As further discussed in the remainder of the paper, this property is of paramount importance whenever the inner operation of an automatic predictor must be interpreted and understood by a human agent.

\paragraph{Neural networks}
\label{par:nn}

Neural networks are biologically-inspired computational models, made of several elementary units (neurons) interconnected into a graph (commonly, \emph{directed} and \emph{acyclic}, a.k.a.\ DAG) via \emph{weighted} synapses.
Accordingly, the most relevant aspects of NN concern the inner operation of neurons and the particular architecture of their interconnection.

Neurons are very simple numeric computational units.
They accept $n$ scalar inputs $(x_1, \ldots, x_n) = \mathbf{x} \in \mathbb{R}^n$ weighted by as many scalar weights $(w_1, \ldots, w_n) = \mathbf{w} \in \mathbb{R}^n$, and they process the linear combination $\mathbf{x} \cdot \mathbf{w}$ via an activation function $\sigma : \mathbb{R} \mapsto \mathbb{R}$, producing a scalar output $y = \sigma(\mathbf{x} \cdot \mathbf{w})$. 
%
%
The output of a neuron may become the input of many others, possibly forming \emph{networks} of neurons having arbitrary topologies.
These networks may be fed with any numeric information encoded as vectors of real numbers by simply letting a number of neurons produce constant outputs.

While virtually all topologies are admissible for NN, not all are convenient.
Many convenient \emph{architectures} -- roughly, patterns of well-studied topologies -- have been proposed in the literature \citep{VanVeenL2019} to serve disparate purposes---far beyond the scope of supervised machine learning.
However, identification of the most appropriate architecture for any given task is non-trivial: recent efforts propose to learn their construction automatically~\citep{liu_progressive_2018,gnn2gnn-uai2022}.
%
%


Most common NN architectures are feed-forward, meaning that neurons are organised in \emph{layers}, where neurons from layer $i$ can only accept ingoing synapses from neurons of layers $j < i$.
The first layer is considered the input layer, which is used to \emph{feed} the whole network, while the last one is the output layer, where predictions are drawn.
In NN architectures inference lets information flow from the input to the output layer -- assuming the weights of synapses are fixed --, while training lets information flow from the output to the input layer---causing the variation of weights to minimise the prediction error of the overall network.

The recent success of deep learning \citep{GoodfellowBC2016} has proved the flexibility and the predictive performance of \emph{deep} neural networks (DNN).
`Deep' here refers to the large amount of (possibly \emph{convolutional}) layers.
In other words, DNN can learn how to apply cascades of convolutional operations to the input data.
Convolutions let the network spot relevant features into the input data, at possibly different scales.
This is why DNN are good at solving complex pattern-recognition tasks---e.g., computer vision or speech recognition.
Unfortunately, however, unprecedented predictive performances of DNN come at the cost of their increased internal complexity, non-inspectability, and greater data greediness.

\subsubsection{General Supervised Learning Workflow}

Briefly speaking, an ML workflow is the process of producing a suitable predictor for the available data and the learning task at hand, with the purpose of exploiting the predictor later so as to draw analyses or to drive decisions.
Hence, any ML workflow is commonly described as composed of two major phases, namely training -- where predictors are fitted on data -- and inference---where predictors are exploited.
However, in practice, further phases are included, such as data provisioning and pre-processing, as well as model selection and assessment.

In other words, before using a sub-symbolic predictor in a real-world scenario, data scientists must ensure it has been sufficiently trained and its predictive performance is sufficiently high.
In turn, training requires
\begin{inlinelist}
    \item an adequate amount of data to be available,
    \item a family of predictors to be chosen (e.g., NN, $K$-nearest neighbours, linear models, etc.),
    \item any structural hyper-parameter to be defined (e.g., amount, type, size of layers, $K$, maximum order of the polynomials, etc.),
    \item any other learning-parameter to be fixed (e.g., learning rate, momentum, batch size, epoch limit, etc.).
\end{inlinelist}
Data must therefore be provisioned before training, and, possibly, pre-processed to ease training itself---e.g., by normalising data or by encoding non-numeric features into numeric form.
The structure of the network must be defined in terms of (roughly) input, hidden, and output layers, as well as their activation functions.
Finally, hyper-parameters must be carefully tuned according to the data scientist's experience, and the time constraints and computational resources at hand.
%

Thus, from a coarse-grained perspective, an ML workflow can be conceived as composed of six major phases, enumerated below:
\begin{enumerate}
    \item \textbf{sub-symbolic data gathering:} the first actual step of any ML workflow, where data is loaded in memory for later processing;

    \item \textbf{pre-processing:} the application of several bulk operations to the training data, following several purposes, such as:
    \begin{inlinelist}
        \item homogenise the variation ranges of the many features sampled by the dataset,
        \item detect irrelevant features and remove them,
        \item construct relevant features by combining the existing ones, or
        \item encoding non-numeric features into numeric form;
    \end{inlinelist}

    \item \textbf{predictor selection:} a principled search for the most adequate sort of predictor to tackle the data and the learning task at hand.
    This is where hyper-parameters are commonly fixed;

    \item \textbf{training:} the actual tuning of the selected predictor(s) on the available data.
    This is where parameters are commonly fixed;

    \item \textbf{validation:} measuring the predictive performance of trained predictors, with the purpose of assessing if and to what extent it will generalise to new, unseen data;

    \item \textbf{inference:} the final phase, where trained predictors are used to draw predictions on unknown data---i.e., different data w.r.t.\ the one used for training.
\end{enumerate}

\subsection{Computational Logic}
\label{subsec:sym-ai}

Symbolic KR has always been regarded as a key issue since the early days of AI, as no intelligence can exist without knowledge, and no computation can occur in lack of representation.
When compared to arrays of numbers, symbolic KR is far more flexible and expressive, and, in particular, more intelligible---both machine- and human-interpretable.
Historically, most KR formalisms and technologies have been designed on top of \emph{computational logic} \citep{lloyd1990computational}, that is, the exploitation of formal logic in computer science.
Consider, for instance, \emph{deductive databases} \citep{green1968}, \emph{description logics} \citep{baader2002}, \emph{ontologies} \citep{cimiano2006-ontologies}, \emph{Horn} logic \citep{Mcnulty1977}, \emph{higher-order} logic \citep{VanBenthem2001}, just to name a few.

\subsubsection{Formal Logics}
\label{subsubsec:formal-logics}

Many kinds of logic-based KR systems have been proposed over the years, mostly relying on \emph{first-order logic} (FOL) -- either by restricting or extending it --, e.g., on description logics and modal logics, which have been used to represent, for instance, terminological knowledge and time-dependent or subjective knowledge.
Here, we briefly recall the state of the art of FOL and its most relevant subsets.

\paragraph{First-order logic}
\label{par:fol}

FOL is a general-purpose logic which can be used to represent knowledge symbolically, in a very flexible way.
More precisely, it allows both human and computational agents to express (i.e., write) the properties of -- and the relations among -- a set of entities constituting the \emph{domain of the discourse}, via one or more \emph{formul\ae}---and, possibly, to reason over such formul\ae{} by drawing inferences.
There, the domain of the discourse $\mathbb{D}$ 
is the set of all relevant entities which should be represented in FOL to be amenable of formal treatment, in a particular scenario.
%

Informally, the syntax for the general FOL formula is defined over the assumption that there exist:
\begin{inlinelist}
    \item a set of \emph{constant} or \emph{function} symbols, 
    \item a set of \emph{predicate symbols}, 
    and
    \item a set of \emph{variables}. 
\end{inlinelist}
Under such assumption, a FOL formula is any expression composed of a list of quantified variables, followed by a number of \emph{literals}, i.e., \emph{predicates} that may or may not be prefixed by the negation operator ($\lnot$). 
Literals are commonly combined into expressions via \emph{logic connectives}, such as conjunction ($\wedge$), disjunction ($\vee$), implication ($\rightarrow$), or equivalence ($\leftrightarrow$).
%

Each predicate consists of a predicate symbol, possibly applied to one or more \emph{terms}.
Terms may be of three sorts, namely \emph{constants}, \emph{functions}, or \emph{variables}.
%
%
Constants represent entities from the domain of the discourse.
In particular, each constant references a different entity.
Functions are combinations of one or more entities via a \emph{function symbol}. 
Similarly to predicates, functions may carry one or more terms.
Being containers of terms, functions enable the creation of arbitrarily complex data structures combining several 
elementary terms into composite ones.
Such kind of composability by recursion is what makes the aforementioned definition of `symbolic' valid for FOL.
Finally, variables are placeholders for unknown terms---i.e., for either individual or groups of entities.

Predicates and terms are very flexible tools to represent knowledge.
While terms can be used to represent or reference either entities or groups of entities from the domain of the discourse, predicates can be used to represent relations among entities, or the properties of each single entity.


\paragraph{Intensional vs.\ extensional}

In logic, one may define concepts -- i.e., describe data -- either \emph{extensionally} or \emph{intensionally}.
Extensional definitions are \emph{direct} representations of data.
In the particular case of FOL, this implies defining a relation or set by explicitly mentioning the entities it involves.
Conversely, intensional definitions are \emph{indirect} representations of data.
In the particular case of FOL, this implies defining a relation or set by describing its elements via other relations or sets.
Recursive intensional predicates are very expressive and powerful, as they enable the description of infinite sets via a finite (and commonly small) amount of formul\ae{}---and this is one of the key benefits of FOL as a means for KR.

\subsubsection{Expressiveness vs.\ Tractability: Notable Subsets of FOL}

Tractability deals with the theoretical questions: `can a logic reasoner compute whether a logic formula is true (or not) in \emph{reasonable} time?'.
Such aspects are deeply entangled with the particular reasoner of choice.
Depending on which and how many features a logic includes, it may be more or less \emph{expressive}.
The higher the expressiveness, the more the complexity of the problems which may be represented via logic and processed via inference increases.
This opens to the possibility, for the solver, to meet queries which cannot be answered in practical time, or by relying upon a limited amount of memory---or just cannot get an answer at all.
Roughly speaking, more expressive logic languages make it easier for human beings to describe a particular domain -- usually, requiring them to write less and more concise clauses --, at the expense of a higher difficulty for software agents to draw inferences autonomously---because of computational tractability.
This is a well-understood phenomenon in both computer science and computational logic \citep{LevesqueB87, BrachmanL2004}, often referred to as the \emph{expressiveness/tractability} trade-off.


FOL, in particular, is considered very expressive.
Indeed, it comes with many undecidable, semi-decidable, or simply intractable properties.
Hence, several relevant subsets of FOL have been identified into the literature, often sacrificing expressiveness for tractability.
%
%
Major notions concerning these logics are recalled below.

\paragraph{Horn logic} 

Horn logic is a notable subset of FOL, characterised by a good trade-off among theoretical expressiveness and practical tractability \citep{Makowsky1987}.

\label{par:horn-logic}
Horn logic is designed around the notion of \emph{Horn clause} \citep{Horn1951}.
Horn clauses are FOL formul\ae{} having no quantifiers, and consisting of a disjunction of predicates, where only at most one literal is non-negated---or, equivalently, an implication having a single predicate as post-condition and a conjunction of predicates as pre-condition:
$
    h \leftarrow b_1,\ \ldots,\ b_n
$.
There, $\leftarrow$ denotes logic implication from right to left, commas denote logic conjunction, and all $b_i$, as well as $h$, are predicates of arbitrary arity, possibly carrying FOL terms of any sort---i.e., variables, constants, or functions.
Horn clauses are thus if-then rules written in reverse order, and only supporting conjunctions of predicates as pre-conditions.


Essentially, Horn logic is a very restricted subset of FOL where:
\begin{inlinelist}
    \item formul\ae{} are reduced to clauses, as they can only contain predicates, conjunctions, and a single implication operator, therefore
    \item operators such as $\vee$, $\leftrightarrow$, or $\lnot$ cannot be used,
    \item variables are implicitly quantified, and
    \item terms work as in FOL.
\end{inlinelist}


\paragraph{Datalog}

Datalog is a restricted subset of FOL \citep{datalog}, representing knowledge via function-free Horn clauses---defined in the previous paragraph.
%
%
So, essentially, Datalog is a subset of Horn logic where structured terms (i.e., recursive data structures) are forbidden.
This is a direct consequence of the lack of function symbols.

Similarly to Horn logic, Datalogs's knowledge bases consist of sets of function-free Horn clauses.

\paragraph{Description logics (DL)} 
\label{par:dl}

Description logics are a family of subsets of FOL, generally involving some or no quantifiers, no structured terms, and no $n$-ary predicates such that $n \geq 3$.
In other words, description logics represent knowledge by only leveraging on constants and variables, other than atomic, unary, and binary predicates.

Differences among specific variants of DL lay in which and how many logic connectives are supported, other than, of course, whether negation is supported or not.
The wide variety of DL is due to the well known expressiveness/tractability trade-off.
However, depending on the particular situation at hand, one may either prefer a more expressive ($\approx$ feature rich) DL variant at the price of a reduced tractability (or even decidability) of the algorithms aimed at manipulating knowledge represented through that DL, or \emph{vice versa}.

Regardless of the particular DL variant of choice, it is common practice in the scope of DL to call
\begin{inlinelist}
    \item constant terms, as `individuals' -- as each constant references a single entity from a given domain --,
    \item unary predicates, e.g., as either `classes' or `concepts' -- as each predicate \emph{groups} a set of individuals, i.e., all those individuals for which the predicate is true --,
    \item binary predicates, e.g., as either `properties' or `roles'---as each predicate \emph{relates} two sets of individuals.
\end{inlinelist}
Following such a nomenclature, any piece of knowledge can be represented in DL by tagging each relevant entity with some constant (e.g., an URL), and by defining concepts and properties accordingly.

Notably, binary predicates are of particular interest as they support connecting couples of entities altogether.
This is commonly achieved via subject-predicate-object \emph{triplets}, i.e., ground binary predicates of the form $\langle \functor{a}\ \predication{f}\ \functor{b} \rangle$ -- or, alternatively, $\predication{f}(\functor{a}, \functor{b})$ --, where $\functor{a}$ is the subject, $\predication{f}$ is the predicate, and $\functor{b}$ is the object.
Such triplets allow users to \emph{extensionally} describe knowledge in a readable, machine-interpretable, and tractable way.

Collections of triplets constitute the so-called \emph{knowledge graphs} (KG), i.e., directed graphs where vertices represent individuals, while arcs represent the binary properties connecting these individuals.
These may explicitly or implicitly instantiate a particular \emph{ontology}, i.e., a formal description of classes characterising a given domain, and of their relations (inclusion, exclusion, intersection, equivalence, etc.), as well as the properties they must (or must not) include.

\paragraph{Propositional logic} 

Propositional logic is a very restricted subset of FOL, where quantifiers, terms, and non-atomic predicates are missing.
Hence, propositional formul\ae{} simply consist of expressions involving one or many 0-ary predicates -- i.e., \emph{propositions} --, possibly interconnected by ordinary logic connectives.
There, each proposition may be interpreted as a Boolean variable -- which can either be true or false --, and the truth of formul\ae{} can be computed as in the Boolean algebra.
So, for instance, a notable example of propositional formula could be as follows:
$
    p \wedge \neg q \rightarrow r
$
where $p$ may be the proposition `it is raining', $q$ may be the proposition `there is a roof', whereas $r$ may be the proposition `the floor is wet'.

The expressiveness of propositional logic is far lower than the one of FOL.
For instance, because of the lack of quantifiers, each relevant aspect/event should be explicitly modelled as a proposition.
Furthermore, because of the lack of terms, entities from a given domain cannot be explicitly referenced.
Such lack of expressiveness, however, implies computing the \emph{satisfiability} of a propositional formula is a \emph{decidable} problem---which may be a desirable property in some application scenarios.

Despite propositional logic may appear too trivial to handle common decision tasks where non-binary data is involved, it turns out a number of apparently complex situations can indeed be reduced to a propositional setting.
This is the case for instance of any expression involving numeric variables or constants, arithmetical comparison operators, logic connectives, and nothing more than that.
%
%
In fact, formul\ae{} containing comparisons among variables or constants (or among each others) can be reduced to propositional logic by mapping each comparison into a proposition.

\subsection{eXplainable Artificial Intelligence}

Modern intelligent systems are increasingly adopting \emph{sub-symbolic} predictive models to support their intelligent behaviour.
These are commonly trained following a data-driven approach.
Such wide adoption is unsurprising, given the unprecedented availability of data characterising the last decade.
ML algorithms enable the detection of useful statistical information buried in data, semi-automatically.
Information, in turn, supports decision-making, monitoring, planning, and forecasting virtually in any human activity where data is available.

However, despite its predictive capabilities, ML comes with some drawbacks making it perform poorly in critical use cases.
The most relevant example is algorithmic \emph{opacity}---intuitively, the human struggle to \emph{understand} how ML-based systems operate or attain their decisions.
In particular, we refer to  `opacity' according to the third definition provided by \citeap{Burrell2016}: ``opacity as the way algorithms operate at the scale of application''.
In ML-based applications, complexity -- and therefore opacity -- arises because of the hardly predictable interplay among highly-dimensional datasets, the algorithms processing them, and the way such algorithms may change their behaviour during learning.

Opacity is a serious issue in all those contexts where human beings are liable for their decisions, or when they are expected/required to provide some sort of \emph{explanation} for it---even if the decision has been suggested by software systems.
This may be the case, for instance, in the healthcare, financial, or legal domains.
In such contexts, ML is at the same time both an enabling factor  -- as it automates decision-making -- and a limiting one---as opacity reduces human \emph{control} on decision-making.
The overall effect is general \emph{distrust} w.r.t.\ AI-based solutions.

Opacity is also the reason why ML predictors are called `black boxes' in the literature.
The expression refers to systems where knowledge is not symbolically represented \citep{Lipton2018}.
In absence of symbolic representations, \emph{understanding} the operation of black boxes -- or why they recommend or take particular decisions -- becomes hard for humans.
The inability to understand black-box content and operation may then prevent people from fully trusting (and, therefore, accepting) them.

To make the picture even more complex, current regulations such as the GDPR \citep{gdpr-voigt2017} are starting to recognise the citizens' \emph{right to explanation} \citep{explanation-aimag38}---which eventually mandates \emph{understandability} of intelligent systems.
This step is essential to guarantee algorithmic fairness, to identify potential biases/problems in the training data or in the black box's operation, and to ensure that intelligent systems work as expected.

%
%
\begin{wrapfigure}{r}{.5\linewidth}
    \centering
    \includegraphics[width=\linewidth]{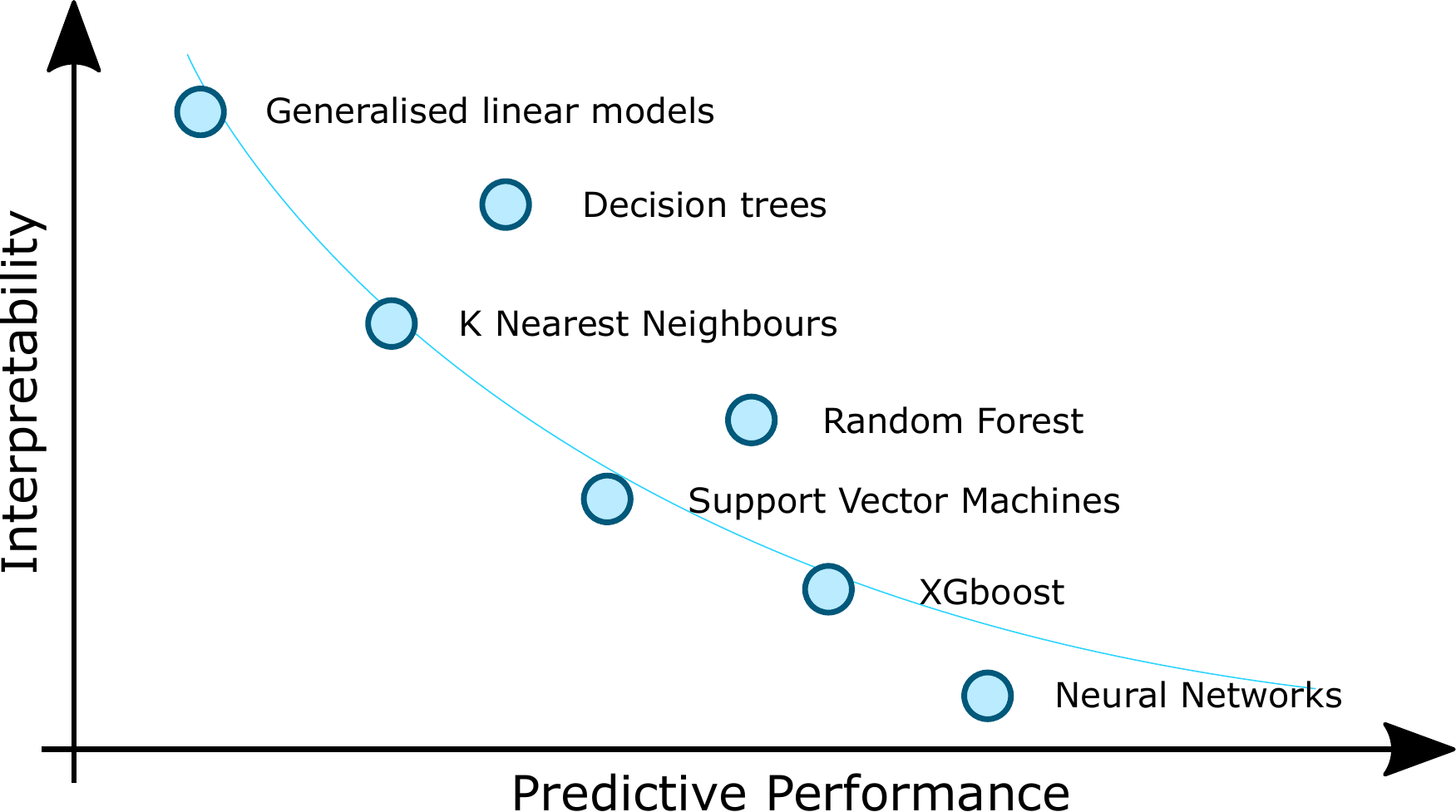}
    \caption{Interpretability/performance trade-off for some common sorts of black-box predictors.}
    \label{fig:tradeoff}
\end{wrapfigure}

Unfortunately, the notion of understandability is neither standardised nor systematically assessed, yet.
No consensus has been reached on what `providing an explanation' should mean when decisions are supported by ML \citep{Miller19}.
However, many authors agree that black boxes are not equally \emph{opaque}: some are more susceptible to interpretation than others for our minds---e.g., \Cref{fig:tradeoff} shows how differences in black-box interpretability are conventionally described.

Despite being informal -- as argued by \cite{Rudin2019}, given the lack of measures for `interpretability' -- \Cref{fig:tradeoff} effectively expresses why research on understandability is actually needed.
Indeed, the figure stresses how the better performing black boxes are also the less interpretable ones.
This is troublesome as, in practice, predictive performance can only rarely be preferred over interpretability.

Nevertheless, consensus has been reached about \emph{interpretability} and \emph{explainability} being desirable properties for intelligent systems.
Hence, within the scope of this paper, we may briefly and informally describe XAI as the corpus of literature and methods aimed at making sub-symbolic AI more interpretable for humans, possibly by automating the production of explanations.

Along this line, based on the preliminary work by \cite{agentbasedxai-aamas2020,agentbasedxai-extraamas2020}, and by drawing inspiration from computational logic (and, in particular, model theory), we let `interpretation' indicate
%
    \textit{``the subjective relation that associates each representation with a specific meaning in the domain of the problem''. }
%
In other words, interpretability refers to the cognitive effort required by human observers to assign a meaning to the way intelligent systems work, or motivate the outcomes they produce.
In those contexts, the notion of interpretability is often coupled with properties as algorithmic transparency (characterising approaches which are \emph{not} opaque), decomposability, or simulatability---i.e., in a nutshell, \emph{predictability}.
Essentially, interpretable systems are understandable when humans can predict their behaviour.

As far as the term \emph{explanation} is concerned, we trace back its meaning to the Aristotelian thought, other than the Oxford dictionary definition, which define `explanation' as
%
    \textit{``a set of statements or accounts that make something clear, or, alternatively, the reasons or justifications given for an action or belief''.}
%
Thus, an explanation is an \emph{activity} aimed at making the relevant details of an object clear or easy to understand to some observer.

Accordingly, the concepts of explainability and interpretability are basically \emph{orthogonal}.
However, they are not unrelated: explanations may consist of constructing better ($\approx$ more-interpretable) representations for the black box at hand.

This is the case, for instance,  of ``explanation by model simplification'' \citep{tolomei2017interpretable}, where a poorly-interpretable model is translated into another -- a more interpretable one --, having ``high fidelity'' \citep{guidotti2018survey} w.r.t.\ the first one.
The translation process of the first model into the second one can be considered as an explanation.
For example, as surveyed by this paper, several methods exist for \emph{extracting} symbolic knowledge out of sub-symbolic predictors.
When this is the case, the extraction act is technically an explanation, as it produces (more) interpretable objects -- the symbolic knowledge -- out of (less) interpretable ones---the predictors.

Conversely, one may regulate the interpretability of an opaque model by altering it to become `consistent' w.r.t.\ (i.e.\ `behave like') some more interpretable one.
In this case, no explanation is involved, yet the resulting model has a higher degree of interpretability---which is commonly the goal.
For instance, as discussed by this paper, several methods exist for \emph{injecting} symbolic knowledge into sub-symbolic predictors.
When this is the case, the injection acts as the means by which opacity issues are worked around.

Interpretability and explainability are key enabling properties for making AI-based solutions (more) trustworthy in the eyes of human users. 
However, as highlighted by \citeap{Rudin2022}, they are not necessarily sufficient: they may enable also \emph{distrust}.
In other words, interpretability and explainability enable finer control on intelligent systems, letting users decide whether to trust them or not.
Along this line, the surveyed SKE/SKI methods should be regarded to as tools for increasing the degree of control users have on AI systems.

\subsubsection{Sorts of Explanation}

According to the main impact surveys in the XAI area \citep{guidotti2018survey, BarredoArrieta2020, xaisurvey-ia14}, two major approaches exist to bring explainability or interpretability features to intelligent systems, namely either \emph{by-design} or \emph{post-hoc}.

\paragraph{XAI by design}

This approach to XAI aims at making intelligent systems interpretable or explainable \emph{ex-ante}, since they are designed to keep these features as first-class goals.
Methods adhering to this approach can be further classified according to two sub-categories:
\begin{description}

    \item[symbols as constraint] containing methods supporting the creation of predictive models -- possibly including or involving some black-box components -- whose behaviour is constrained by a number of symbolic and intelligible rules, usually expressed in terms of (some subset of) FOL.

    \item[transparent box design] containing methods supporting the creation of predictive models that are inherently interpretable, requiring no further manipulation; 

\end{description}
In particular, in the remainder of this paper, we focus on methods from the latter category, as it is deeply entangled with symbolic knowledge \emph{injection}.

\paragraph{Post-hoc explainability}

This approach to XAI aims at making intelligent systems interpretable or explainable \emph{ex-post}, i.e., by somehow manipulating poorly-interpretable pre-existing systems.
Methods adhering to this approach can be further classified according to the following sub-categories:
\begin{description}
    \item[text explanation] where explainability is achieved by generating textual explanations that help to explain the model results; methods that generate symbols representing the model behaviour are also included in this category, as symbols represent the logic of the algorithm through appropriate semantic mapping;
    \item[visual explanation] techniques that allow the visualisation of the model behaviour; several techniques existing in the literature comes along with methods for dimensionality reduction, to make visualisation human-interpretable;
    \item[local explanation] where explainability is achieved by first segmenting the solution space into less complex solution subspaces relevant for the whole model, then producing their explanation;
    \item[explanation by example] allows for the extraction of representative examples that capture the internal relationships and correlations found by the model;
    \item[model simplification] techniques where a completely-new simplified system is built, trying to optimise similarity with the previous one while reducing complexity;
    \item[feature relevance] methods focus on how a model works internally by assigning a relevance score to each of its features, thus revealing their importance for the model in the output.
\end{description}
In particular, in the remainder of this paper, we focus on methods from the `model simplification' category, as it is deeply entangled with symbolic knowledge \emph{extraction}.

\section{Definitions \& Methodology}
\label{sec:preliminaries}

The goal of our SLR is to detect and categorise the many SKE and SKI algorithms proposed into the literature so far, hence shaping a clear picture of what SKE and SKI mean today.

Following this purpose, we
\begin{inlinelist}
    \item\label{step:definitions} start from broad and intuitive definitions of both SKE and SKI (provided in \Cref{sec:ske-ski-definitions}); we then
    \item\label{step:questions} define a number of research questions aimed at delving into the details of actual SKE and SKI methods; along this line, we
    \item explore the literature looking for contributions matching the broad definitions from step \ref{step:definitions} (following a strategy described in \Cref{sec:survey}). Finally, by analysing such contributions, we
    \item provide answers for the research questions from step \ref{step:questions} (in \Cref{sec:results}), and, in doing so, we
    \item synthesise general, bottom-up taxonomies for both SKE and SKI (in \Cref{subsec:ske-taxonomy,subsec:ski-taxonomy}).
\end{inlinelist}

\subsection{Definitions for Symbolic Knowledge Extraction and Injection}
\label{sec:ske-ski-definitions}

Here we provide broad definitions for both symbolic knowledge extraction and injection, following the purpose of drawing a line among what methods, algorithms, and technologies from the literature should be considered related to either SKE or SKI, and what should not.
We do so under a XAI perspective, hence highlighting how both SKE and SKI help mitigating the opacity issues arising in data-driven AI.
Then we discuss the potential arising from the \emph{joint} exploitation of both SKE and SKI.

Notably, we tune our definitions so as to comprehend and generalise the many methods and algorithms surveyed later in this paper.
Indeed, looking for a wider degree of generality, our definitions commit to no particular form of symbolic knowledge, nor sub-symbolic predictor---despite many surveyed techniques come with commitments of that sort.
Hence, in what follows we write `symbolic knowledge' meaning `any chunk of intelligible information expressed in \emph{any} possibly sort of logic', as well as any sort of information which can be rewritten in logic form (e.g., decision trees).
Similarly, we write `sub-symbolic predictor' meaning `any sort of \emph{supervised} ML model which can be fitted over \emph{numeric} data to eagerly solve classification or regression tasks'.

\subsubsection{Extraction}
\label{subsubsec:ske-definition}
Generally speaking, SKE serves the purpose of generating intelligible representations for the sub-symbolic knowledge an ML predictor has grasped from data during learning.
Here we provide a general definition of SKE and discuss its purpose as well as the major benefits it brings against the XAI landscape.

\paragraph{Definition}

We define SKE as
\begin{displayquote}\itshape
    any \emph{algorithmic} procedure accepting \emph{trained} sub-symbolic predictors as input and producing \emph{symbolic} knowledge as output, so that the extracted knowledge reflects the behaviour of the predictor with high \emph{fidelity}.
\end{displayquote}
Notably, this definition emphasises a number of key aspects of SKE which are worth to be described in further detail.

First, SKE is modelled as a class of \emph{algorithms} -- hence finite-step recipes -- characterised by what they accept as input and what they produce as output.

As far as the inputs of SKE procedures are concerned, the only explicit requirement is on \emph{trained} ML predictors.
There is no constraint w.r.t.\ the nature of the predictor itself, hence SKE procedures may be designed for any possible predictor family, in principle.
Yet, this requirement implies that the predictor's training has already occurred, and it has reached some satisfying performance w.r.t.\ the task it has been trained for.
Hence, in an ML workflow, SKE should occur \emph{after} training and validation were concluded.

As far as the outputs of SKE procedures are concerned, the only explicit requirement is about the production of \emph{symbolic} knowledge.
`Symbolic' is here intended, in a broader sense, as a synonym of `intelligible' (for the human being), hence admissible outcomes are logic formul\ae{} as well as decision trees, or bare human-readable text.

In any case, for an algorithm to be considered a valid SKE procedure, the output knowledge should mirror the behaviour of the original predictor w.r.t.\ the domain it was trained for, as much as possible.
This involves some fidelity score aimed at measuring how well the extracted knowledge mimics the predictor it was extracted by, w.r.t.\ the domain and the task that predictor was trained for.
This, in turn, implies that the extracted knowledge should, in principle, act as a predictor as well, thus being queryable as the original predictor would.
Thus, for instance, if the original predictor is an image classifier, the extracted knowledge should let an intelligent agent classify images of the same sort, expecting the same result.
The agent may then be either computational (i.e., a software program) or human, depending on whether the extracted knowledge is machine- or human-interpretable.
Notably, the exploitation of \emph{logic} knowledge as the target of SKE is of particular interest as it would enable both options.

\paragraph{Purpose and benefits}

Generally speaking, one may be interested in performing SKE to inspect the inner operation of an opaque predictor, which should be considered a black box otherwise.
However, one may also perform SKE to automatise and speed up the process of acquiring symbolic knowledge, instead of crafting knowledge bases manually.

Inspecting a black-box predictor through SKE, in turn, is an interesting capability within the scope of XAI.
Given a black-box predictor and a knowledge-extraction procedure applicable to it, any extracted knowledge can be adopted as a basis to construct explanations for that particular predictor.
Indeed, the extracted knowledge may act as an \emph{interpretable replacement} (a.k.a.\ surrogate model) for the original predictor, provided that the two have a high fidelity score~\citep{agentbasedxai-aamas2020}. 

Accordingly, the application of SKE to XAI brings a number of relevant opportunities, e.g., by letting human users
\begin{inlinelist}
    \item study the internal operation of an opaque predictor to find, for instance, mispredicted input patterns; or correctly predicted input patterns leveraging upon some unethical decision process;
    \item highlight the differences or the common behaviours between two or more black-box predictors performing the same task;
    \item merge the knowledge acquired by various predictors -- possibly of different kinds -- on the same domain---provided that the same representation format is used for extraction procedures~\citep{xmas-aiiot2019}.
\end{inlinelist}

\subsubsection{Injection}
\label{subsubsec:ski-definition}
Generally speaking, SKI serves a dual purpose w.r.t.\ to SKE.
In particular, SKI aims at letting an ML predictor keep some symbolic knowledge into account when drawing predictions.
Here, we provide a general definition of SKI, and we discuss its purpose and the major benefits it brings w.r.t.\ the XAI panorama.

\paragraph{Definition}

We define SKI as
\begin{displayquote}\itshape
    any \emph{algorithmic} procedure affecting how sub-symbolic predictors draw their inferences in such a way that predictions are either \emph{computed} as a function of, or made \emph{consistent} with, some \emph{given} symbolic knowledge.
\end{displayquote}

This definition emphasises a number of key aspects of SKI which are worth to be described in further detail.
Similarly to SKE, it is modelled as a class of \emph{algorithms}.
Yet, dually w.r.t.\ extraction, SKI algorithms are procedures accepting symbolic knowledge as input and producing ML predictors as output.

About the inputs of SKI procedures, the only explicit requirement is that knowledge should be symbolic and user-provided---hence \emph{human}-interpretable.
However, since any input knowledge should be algorithmically manipulated by the SKI procedure, we elicit an implicit requirement here, constraining the input knowledge to be \emph{machine}-interpretable as well.
This implies that some formal language -- e.g., some formal logic, or some decision tree -- should be employed for knowledge representation, while free text or natural language should be avoided.

Along this line, another implicit requirement is that the input knowledge should be \emph{functionally analogous} w.r.t.\ the predictors undergoing injection.
In other words, if a predictor aims at classifying customer profiles as either worthy or unworthy for credit, then the symbolic knowledge should encode decision procedures to serve the exact same purpose, and observe the exact same information.

About the outcomes of SKI procedures, our definition identifies two relevant situations---which are not necessarily mutually-exclusive.
On the one side, SKI procedures may enable sub-symbolic predictors to accept symbolic knowledge as input.
SKI procedures of this sort essentially consist of pre-processing algorithm aimed at encoding symbolic knowledge in sub-symbolic form, hence enabling sub-symbolic predictors to accept them as input.
In this sense, SKI procedures of this sort enable sub-symbolic predictors to (learn how to) \emph{compute} predictions as functions of the symbolic knowledge they were fed with---assuming it has been conveniently converted into sub-symbolic form.
On the other side, SKI procedures may alter sub-symbolic predictors so that they draw predictions which are \emph{consistent} with the symbolic knowledge---according to some notion of \emph{consistency}.
SKI procedures of this sort essentially affect either the structure or the training process of the sub-symbolic predictors they are applied to, in such a way that the predictor must then keep the symbolic knowledge into account when drawing predictions.
In this sense, SKI procedures of this sort force sub-symbolic predictors to learn not only from data but from symbolic knowledge as well.

In any case, regardless of their outcomes, SKI procedures fit the ML workflow in its early phases, as they may affect both pre-processing and training.

Notably, consistency plays a pivotal role in SKI, dually w.r.t.\ what fidelity does for SKE.
Along this line, our definition involves some consistency score aimed at measuring how well the predictor undergoing injection can take advantage from the injected knowledge, w.r.t.\ the domain and the task that predictor was trained for.
So, for instance, if a knowledge base states that loans should be guaranteed to people from a given minority -- as long as annual income overcomes a given threshold --, then any predictor undergoing injection of that knowledge base should output predictions respecting that statement---or at least minimise violations w.r.t.\ it.

\paragraph{Purpose and benefits}
Generally speaking, one may be interested in performing SKI to reach a higher degree of control on what a sub-symbolic predictor is learning.
In fact, SKI may either incentivise the predictor to learn some desirable behaviour, or discourage it from learning some undesired behaviour.
However, one may also exploit SKI to perform sub-symbolic or fuzzy manipulations of symbolic knowledge, which would be otherwise unfeasible or hard to formalise via crisp symbols.
While the latter option is further analysed by a number of authors -- such as \cite{gnn-woa2021,LambGGPAV20} --, in the remainder of this section we focus on the former use case, as it is better suited to serve the purposes of XAI.

Within the scope of XAI, SKI is a remarkable capability as it provides a workaround for the issues arising from the opacity of ML predictors.
While SKE aims at reducing the opacity of predictor by letting users understand its behaviour, SKI aims at bypassing the need for transparency.
Indeed, predictors undergoing the injection of \emph{trusted} symbolic knowledge provide higher guarantees about their behaviour, which will be more predictable and comprehensible.

Accordingly, the application of SKI to XAI brings a number of relevant opportunities, e.g., by letting the human designers
\begin{inlinelist}
    \item endow sub-symbolic predictors with their common sense, and, therefore:
    \item finely control what predictors are learning, and, in particular,
    \item let predictors learn about relevant situations, despite poor data is available to describe them.
\end{inlinelist}
Provided that adequate SKI procedures exist, all such use cases come at the price of handcrafting \emph{ad hoc} knowledge bases reifying the designers' common sense in symbols, and then injecting it in ordinary ML predictors.

\subsection{Review Methodology}
\label{sec:survey}


The overall review workflow is inspired by the goal question metric approach by \cite{goal-question-metric}. 
In short, the workflow requires some clear research \emph{goal(s) } to be fixed, and then decomposed into a number of research question the survey will then provide answers to.
To produce such answers, the workflow requires of course scientific papers to be selected, and analysed.
To serve this purpose, the workflow requires a pool of \emph{queries} to be identified.
Such queries must be performed on most relevant bibliographic search engines (e.g., Google Scholar, Scopus).
Finally, the workflow requires the query results to be selected (or excluded) for further analyses following a reproducible criterion.
Any subsequent analysis is then devoted to answer the aforementioned research questions, hence drawing useful classifications and general conclusions.

For the sake of reproducibility, in the remainder of this subsection we delve into the details of how our SLR on symbolic knowledge extraction and injection is conducted.

We start by defining three different research goals (\textbf{G}):
\begin{enumerate}[label=\textbf{G\arabic{*}}]
    \item\label{item:SKE} -- \emph{understanding which are the features of SKE algorithms}
    \item\label{item:SKI} -- \emph{understanding which are the features of SKI algorithms}
    \item\label{item:tech} -- \emph{probing the current level of technological readiness of SKE/SKI technologies}
\end{enumerate}
Then, we break them down in the following research questions (\textbf{RQ}):
\begin{enumerate}[label=\textbf{RQ\arabic{*}}]
    \item\label{item:rq1} (from \ref{item:SKE}) -- \emph{which sort of ML predictors can SKE be applied to?}
    \item\label{item:rq2} (from \ref{item:SKE}) -- \emph{is there any requirement on the input data for SKE?}
    \item\label{item:rq3} (from \ref{item:SKE}) -- \emph{which kind of SK can be extracted from ML predictors?}
    \item\label{item:rq4} (from \ref{item:SKE}) -- \emph{for which kind of AI tasks can SKE be exploited?}
    \item\label{item:rq5} (from \ref{item:SKE}) -- \emph{how does SKE work?}
    \item\label{item:rq6} (from \ref{item:SKI}) -- \emph{which sorts of ML predictors can SKI be applied to?}
    \item\label{item:rq7} (from \ref{item:SKI}) -- \emph{which kind of SK can be injected into ML predictors?}
    \item\label{item:rq8} (from \ref{item:SKI}) -- \emph{for which kind of AI tasks can SKI be exploited?}
    \item\label{item:rq9} (from \ref{item:SKI}) -- \emph{how does SKI work?}
    \item\label{item:rq10} (from \ref{item:tech}) -- \emph{which and how many SKE/SKI algorithms come with runnable software implementations?}
\end{enumerate}
Notice that research questions about SKE are analogous to those about SKI.
In both cases, research questions are devoted to clarify which kind of information can SKE (resp.\ SKI) methods accept as input (resp.\ produce as output), how do they work, which AI tasks they can be used for (e.g., regression, classification), and which ML predictors they can be applied to (e.g., NN, SVM, etc.).

In order to answer the research questions above, we identify a number of queries to be performed on widely-available bibliographic search engines.
In detail, queries involve the following keywords:
\begin{itemize}
    \item (`rule extraction' $\vee$ `knowledge extraction') $\wedge$ (`neural networks' $\vee$ `support vector machines')
    \item (`pedagogical' $\vee$ `decompositional' $\vee$ `eclectic') $\wedge$ (`rule extraction' $\vee$ `knowledge extraction')
    \item `symbolic knowledge' $\wedge$ (`deep learning' $\vee$  `machine learning')
    \item `embedding' $\wedge$ (`knowledge graphs' $\vee$  `logic rules' $\vee$ `symbolic knowledge')
    \item `neural' $\wedge$ `inductive logic programming'
\end{itemize}
As far as bibliographic search engines are concerned, we exploit Google Scholar,\footnote{\url{https://scholar.google.com}} Scopus,\footnote{\url{https://www.scopus.com}} Springer Link,\footnote{\url{https://link.springer.com}} ACM Digital Library,\footnote{\url{https://dl.acm.org}} and DBLP.\footnote{\url{https://dblp.uni-trier.de}}

For each search engine and query pair, we consider the first two pages of results.
For each result, we inspect the title, abstract, and -- in case of ambiguity --, the introduction, while trying and classifying it according to three disjoint circumstances:
\begin{inlinelist}
    \item the paper is a \emph{primary work} describing some SKE or SKI method matching the broad definitions from \ref{sec:ske-ski-definitions},
    \item\label{step:secondary} the paper is a \emph{secondary work} surveying some portion of literature overlapping SKE or SKI (or both),
    \item the paper is \emph{unrelated} w.r.t.\ to both SKE and SKI, hence it is not relevant for this survey.
\end{inlinelist}
Notably, secondary works selected in step \ref{step:secondary} are valuable sources of primary works, hence we recursively explored their bibliographies to further select other primary works.
In particular, in this phase we leverage upon relevant secondary works such as \cite{andrews1995survey,BesolddABBDHKLLMvlPPPZ2017,xaisurvey-ia14,WangMWG17,GarcezBG01,guidotti2018survey,Hailesilassie16,Huysmans2006UsingRE,vonRuedenMBGGHKWPPRGBS2021,XieXMKS19,ZilkeMJ16}---which we acknowledge as noteworthy (even though less extensive) surveys in the field of SKE or SKI.

We select \allcount{} primary works, of which \skecount{} works concern SKE, and \skicount{} concern SKI.
We then analyse each primary work individually, in order to provide answers to the aforementioned research questions.
While doing so, we construct bottom-up taxonomies for both SKE and SKI.

Finally, we inspect each primary work for assessing its technological status.
In particular, we look for runnable software implementations corresponding to the method described in the primary work.
In case no software tool is clearly mentioned in the primary work, or if the software is not technically accessible (e.g., Web site or repository is private or non-reachable) at the time of the survey, then we consider the method as lacking software implementations.
Otherwise, we further distinguish among methods coming with reusable software libraries, and methods coming with experimental code.
In the first case, the software is ready for re-use, either because it is published on public software repositories such as PyPi, or because it is structured in such a way to let users exploit it for custom purposes.
Vice versa, if the software is tailored on the experiments mentioned in the primary work, then we consider it experimental.

\section{Survey Results}
\label{sec:results}

This section summarises the results of our survey.
In particular, answers for the research questions outlined in \Cref{sec:survey} are provided here.

Accordingly, we group research questions according to their main focus (SKE or SKI), and we answer to each question individually---grouping answers when convenient, for the sake of conciseness.
Answers consist of brief statistical reports showing the distribution of the surveyed SKE/SKI methods w.r.t.\ some dimension of interest for either SKE or SKI.
Interesting dimensions are presented on the fly, as part of our answers.
This is deliberate, since we select as `interesting dimension' any relevant way of clustering the surveyed methods.
In other words, we let taxonomies emerge from the literature rather than super-imposing any particular view of ours.

\subsection{Symbolic Knowledge Extraction}
\label{subsec:ske-taxonomy}

By building upon secondary works, such as the work by \cite{xaisurvey-ia14} and the survey of \cite{andrews1995survey}, we identify three relevant dimensions by which SKE methods can be categorised, namely:
\begin{inlinelist}
    \item the learning task(s) they support;
    \item the method's translucency;
    \item the shape of the extracted knowledge.
    By analysing the surveyed SKE methods, we find these categories adequate.
    However, we identify new dimensions, namely:
    \item the sort of input data the predictor undergoing extraction is trained upon, and
    \item the expressiveness of the extracted knowledge.
\end{inlinelist}
In what follows we answer research questions \ref{item:rq1}--\ref{item:rq5} and \ref{item:rq10} by focusing on such dimensions, individually.
Conversely, in the supplementary materials (see \Cref{app:summary-ske}), we provide an overview of the \skecount{} methods selected for SKE.

\subsubsection{\ref{item:rq1}: Which sort of ML predictors can SKE be applied to? \ref{item:rq5}: How does SKE work?}
\label{subsubsec:RQ1}

Answers for questions \ref{item:rq1} and \ref{item:rq5} are deeply entangled, as they are both related to SKE methods' \emph{translucency}.
Translucency deals with the need of SKE methods to inspect the internal structure of the underlying black-box model, while producing the extracted rules.

SKE methods provide for translucency in two ways \cite{andrews1995survey}, and can be labelled accordingly as
\begin{description}
    \item[decompositional] 
    if the method needs to inspect (even partially) the internal parameters of the underlying black-box predictor, e.g., neuron biases or connection weights for NN, or support vectors for SVM;
    \item[pedagogical] 
    if the algorithm \emph{does not} need to take into account any internal parameter, but it can extract symbolic knowledge by only relying on the predictor's outputs.
\end{description}
Along this line, we observe that surveyed SKE methods can be grouped into as many big clusters, depending on how they treat the predictor undergoing extraction.

\begin{SCfigure}
    \centering
	\includegraphics[width=.45\linewidth]{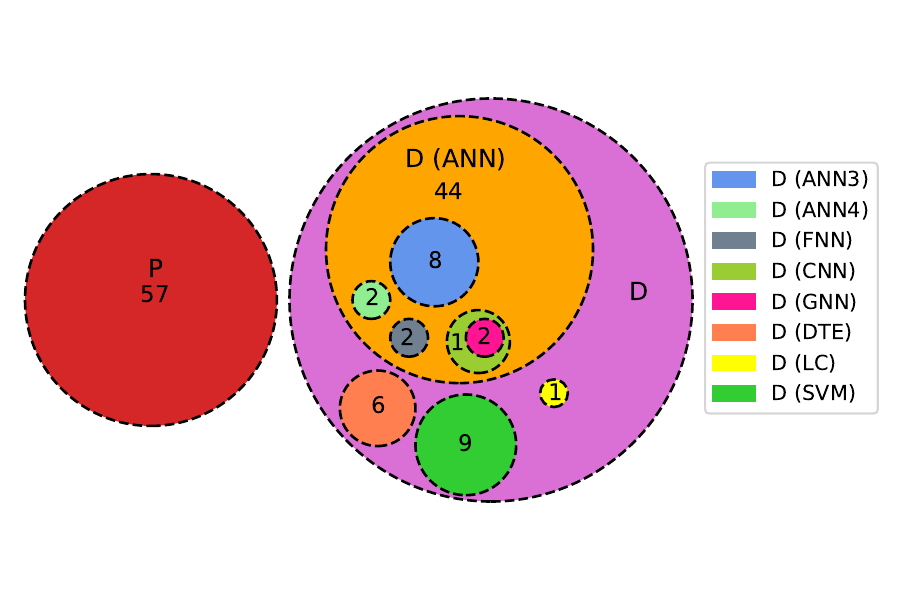}
    \caption{
        Venn diagram categorising SKE methods w.r.t.\ \emph{translucency}: pedagogical (P) or decompositional (D).
        For decompositional methods, we report the target predictor type: ANN$\langle n \rangle$ = artificial NN (possibly, having exactly $\langle n \rangle$ layers), CNN = convolutional NN, GNN = graph NN, FNN = fuzzy NN, SVM = support vector machines, DTE = decision tree ensembles, LC = linear classifiers.
    }
    \label{fig:pie-ske-trans}
\end{SCfigure}

W.r.t.\ \ref{item:rq1}, it is worth highlighting that pedagogical methods can be applied to any sort of supervised ML predictor, in principle---despite the literature may only report particular cases of application to specific predictors.
Conversely, each decompositional method focuses on a specific sort of supervised ML predictor.
Hence, decompositional SKE methods can be further categorised w.r.t.\ which sort of supervised ML predictors they are tailored upon.
As detailed by \Cref{fig:pie-ske-trans}, the translucency is far from uniform for SKE methods.
Indeed, nearly a half of the surveyed methods are pedagogical, while the rest are tailored on feed-forward NN (possibly, with fixed amounts of layers), SVM, linear classifiers, or decision tree ensembles.


W.r.t.\ \ref{item:rq5}, it is worth highlighting that pedagogical methods treat the underlying predictor as an \emph{oracle}, to be queried for predictions the symbolic knowledge shall emulate.
Conversely, decompositional methods must look into the internal structure of predictors, hoping to detect meaningful patterns.
For instance, SKE methods focusing on NN may try to interpret inner neurons as meaningful expressions combining their ingoing synapses.

\subsubsection{\ref{item:rq2}: Is there any requirement on the input data for SKE?}
\label{subsubsec:RQ2}

\begin{SCfigure}
    \centering
    \includegraphics[width=.4\linewidth]{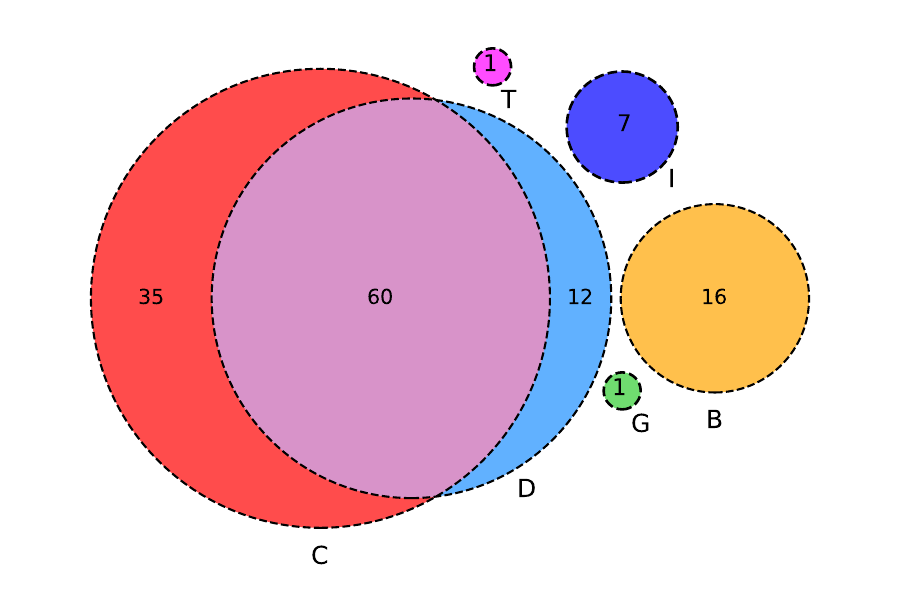}
    \caption{Venn diagram categorising SKE methods w.r.t.\ the \emph{input data type} required by the underlying predictor: binary (B), discrete (D), continuous (C), images (I), text (T), graphs (G).}
    \label{fig:pie-ske-input}
\end{SCfigure}

This question can be answered by looking at the accepted input data type of the surveyed SKE methods.
In most cases data is structured, i.e., it consists of tables of numbers, where features are of three different sorts:
\begin{description}
    \item[binary] if the feature can assume only two values, generally encoded with 0 and 1 (or -1 and 1, or \pl{true} and \pl{false});

    \item[discrete] if the feature can assume values drawn from a \emph{finite} set of admissible values;
    notably, when this is the case, data science identifies two relevant sub-sorts of features:
    %
        \textbf{ordinal} if the set of admissible values is \emph{ordered} (hence, enabling the representation of the feature via some range of integer numbers), or
        \textbf{categorical} if that set is \emph{un}ordered (hence, enabling the representation of the feature via one-hot encoding);

    \item[continuous] if the feature can assume any real numeric value.
\end{description}
Alternatively, data may consist of
\begin{description}
    \item[images] i.e., matrices of pixels, possibly with multiple channels;
    \item[text] i.e., sequences of characters of arbitrary length;
    \item[graphs] i.e., data structures of variable size, consisting of nodes/vertices interconnected by edges/arcs.
\end{description}
In \Cref{fig:pie-ske-input}, we report absolute occurrence of the sorts of input features accepted by the surveyed SKE methods, as described by their authors.
%

As the reader may notice, the vast majority of surveyed methods are tailored on structured data with \emph{continuous} and/or \emph{discrete} features.

\subsubsection{\ref{item:rq3}: Which kind of SK can be extracted from ML predictors?}
\label{subsubsec:RQ3}

Broadly speaking, any extracted SK should mirror (i.e., mimic) the operation of the ML predictor it has been extracted from.
For \emph{supervised} ML, this means the extracted knowledge should express a \emph{function}, mapping input features into output features (e.g.\ classes, for classification tasks).
Functions can be represented in symbols in several ways.
Indeed, the SK extracted by the surveyed methods comes in various form.

Notably, such forms can be categorised under both a \emph{syntactic} or \emph{semantic} perspective.
There, syntax refers to the \emph{shape} of the extracted SK, whereas semantic refers to what kind of logic formalism the extracted knowledge may leverage upon---which is a matter of \emph{expressiveness}.

\paragraph{Shape of the extracted knowledge}


\begin{SCfigure}
    \centering
    \includegraphics[width=.35\linewidth]{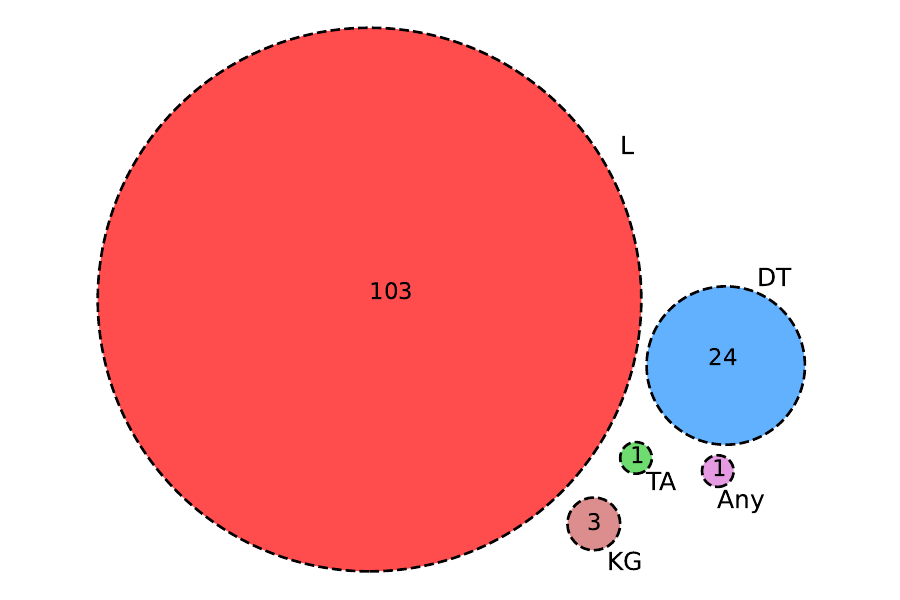}
    \caption{Venn diagram categorising SKE methods w.r.t.\ the output knowledge's \emph{shape}: rule lists (L), decision trees (DT) or tables (TA), knowledge graphs (KG).}
    \label{fig:pie-ske-rule-shape}
\end{SCfigure}

As far as syntax is concerned, decision \emph{rules} \citep{freitas2014comprehensible,huysmans2011empirical,murphy1991id2} and \emph{trees} \citep{breiman1984classification,quinlan1993c4} are the most widespread human-comprehensible formats for the output knowledge, thus the vast majority of surveyed methods adopt one of these.
However, other solutions have been exploited as well---e.g., decision \emph{tables}.
In all cases, however, a common trait is that functions of real numbers are expressed by using \emph{symbols} to denote the same input and output features the underlying ML predictor was trained upon.

W.r.t.\ surveyed SKE methods, we identify four major admissible shapes:
\begin{description}
    \item[lists] of rules, i.e.\ sequences of logic rules to be read in some predefined order;

    \item[decision trees] see \Cref{par:decision-trees};

    \item[decision tables] i.e., concise visual rule representations specifying one or more conclusions for each set of different conditions.
    They can be \emph{exhaustive} -- if all the possible combinations are listed --, or \emph{incomplete}---otherwise.
    Generally speaking, decision tables are structured as follows: there is a column (row) for each input and output variable and a row (column) for each rule.
    Each cell $c_{ij}$ ($c_{ji}$) contains the value of the $j$-th variable for the $i$-th rule.
    An example of decision table is provided in the supplementary material (see \Cref{app:examples})

    \item[knowledge graphs] see \Cref{par:dl}.
\end{description}
\Cref{fig:pie-ske-rule-shape} sums up the occurrence of the different shapes of output rules required for SKE algorithms.
As the reader may notice, the majority of the surveyed methods target rule \emph{lists}.
Arguably, this trend may be motivated by the great simplicity of rule lists, in terms of readability, and their algorithmic tractability.

\paragraph{Expressiveness of the extracted knowledge}

Despite the extracted knowledge may contain statements of different shapes (e.g., rules, trees, tables), the readability, conciseness, and tractability of the extracted rules heavily depend on what can those statements contain---which, in turn, dictate what can (or cannot) be expressed.
In the general case, statements may contain \emph{predicates} or \emph{relations} among the symbols representing input or output features.
These may (or may not) contain logic connectives as well as arithmetic or logic comparators.
SKE methods can be categorised w.r.t.\ which and how many ways of combining symbols are admissible within statements.

Along this line, we identify five major formats for statements in the surveyed SKE methods:
\begin{description}
    \item[propositional rules] are the simplest format, where statements consist of \emph{propositions} -- i.e.\ symbols denoting \emph{boolean} input/output features --, possibly interconnected via logic connectives (negation, conjunction, disjunction, etc.).
    Notice that statements containing relations (e.g., arithmetic comparisons) among \emph{single}, \emph{continuous} features and \emph{constant} values are indeed propositional as well.

    \item[fuzzy rules] are propositional rules where the truth value of conditions and conclusions are not limited to 0 and 1, but can assume any value $\in [0,~1]$;
    
    \item[oblique rules] have conditions expressed as inequalities involving linear combinations of the input variables.
    This is different from the propositional case, as features may be compared to other features (rather than constants alone).

    \item[\xofy{m}{n} rules] are particular sorts of rules where \emph{boolean} statements are grouped by $n$ and each rule is \emph{true} only if at least $m$ literals (out of $n$) are \emph{true}, with $m \leq n$.
    Notice that \xofy{m}{(X_1, \ldots, X_n)} is just a concise way of writing the disjunction among the conjunction of all possible $m$-sized combinations of $n$ boolean literals $X_1, \ldots, X_n$.
    Hence, \xofy{m}{n} rules are just a concise way of writing rules of other sorts: if $X_1, \ldots, X_n$ are all predicative statements, then the expression \xofy{m}{(X_1, \ldots, X_n)} is predicative as well---and the same is true if $X_1, \ldots, X_n$ are oblique statements.
    
    \item[triplets] see \Cref{par:dl}.
\end{description}
%

\begin{SCfigure}
    \centering
    \includegraphics[width=.45\linewidth]{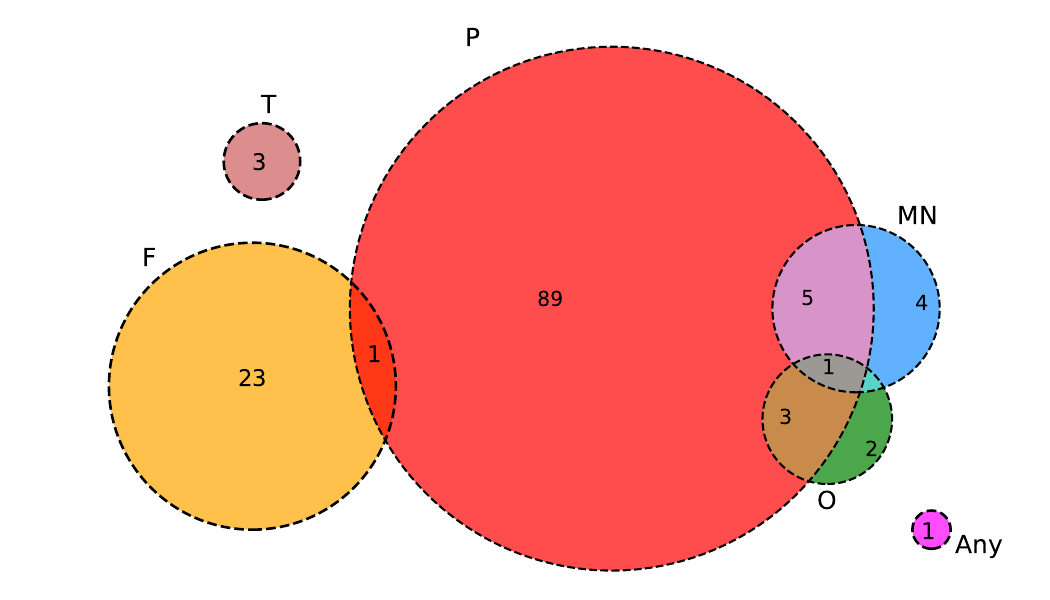}
    \caption{Venn diagram categorising SKE methods w.r.t.\ the output knowledge's \emph{expressiveness}: propositional (P), \mofn{} (MN), fuzzy (F), or oblique (O) rules; or triplets (T).}
    \label{fig:pie-ske-rule-format}
\end{SCfigure}

\Cref{fig:pie-ske-rule-format} summarises the occurrence of the different SK formats produced by the surveyed SKE algorithms.
As the reader may notice, the vast majority of surveyed SKE methods produce predicative rules, i.e.\ rules composed of several boolean statements about individual input features, possibly interconnected via logic connectives.
Arguably, this trend may be motivated by the great tractability of propositional rules, and by their simplicity.
In fact, to construct propositional rules, SKE algorithms may follow a \emph{divide-et-impera} approach by focusing on each single input feature at a time---hence enabling the simplification of the extraction process itself.

\subsubsection{\ref{item:rq4}: For which kind of AI tasks can SKE be exploited?}
\label{subsubsec:RQ4}

ML methods are commonly exploited in AI to serve specific purposes, e.g.\ classification, regression, clustering, etc.
Regardless of the particular means by which SKE is attained, extraction aids the human users willing to inspect \emph{how} those methods work.
However, the particular AI tasks ML predictors have been designed for play a pivotal role in determining what outputs users may expect from those predictors.
A similar argument holds for extraction procedures, as the extracted knowledge should reflect the inner behaviour of the original predictor.
Along this line, it is interesting to categorise SKE methods w.r.t.\ the AI task they assume for the ML predictors they are applied to.

\begin{SCfigure}
    \centering
    \includegraphics[width=.3\linewidth]{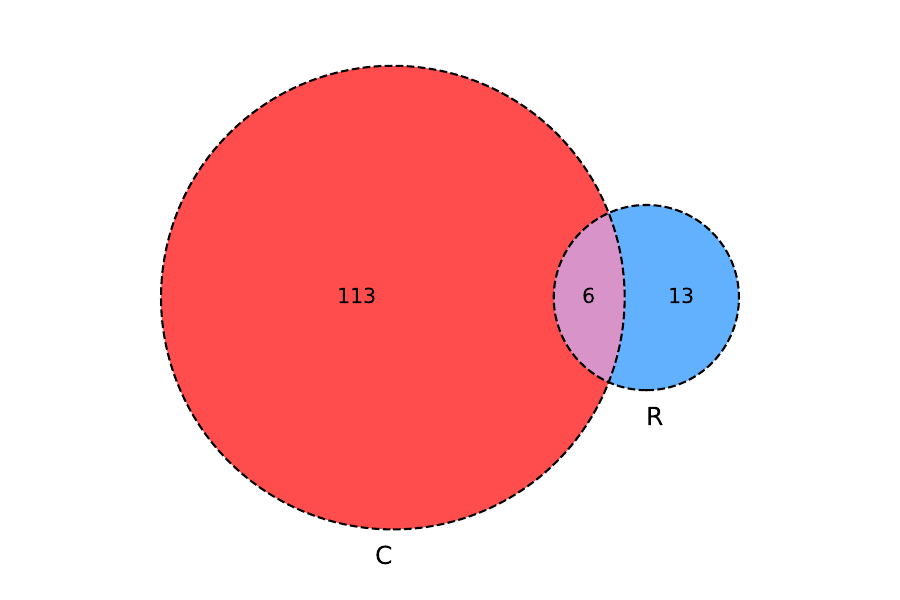}
    \caption{Venn diagram categorising SKE methods w.r.t.\ the \emph{targeted AI task}: classification (C) or regression (R).}
    \label{fig:pie-ske-task}
\end{SCfigure}

\Cref{fig:pie-ske-task} summarises the occurrence of tasks among the surveyed SKE methods.
Notably, most of them can be applied uniquely to \emph{classifiers}, whereas a small portion of them is explicitly designed for \emph{regressors}.
Only few methods can handle both categories.

In general, we observe how the surveyed methods are tailored on either classification or regression tasks---when not both.
In either cases, surveyed methods focus on supervised ML tasks.
To the best of our knowledge, currently, there are no SKE procedures tailored on unsupervised or reinforcement learning tasks.


\subsubsection{\ref{item:rq10}: which and how many SKE algorithms come with runnable software implementations?}


\begin{SCfigure}
    \centering
	\includegraphics[width=.35\linewidth]{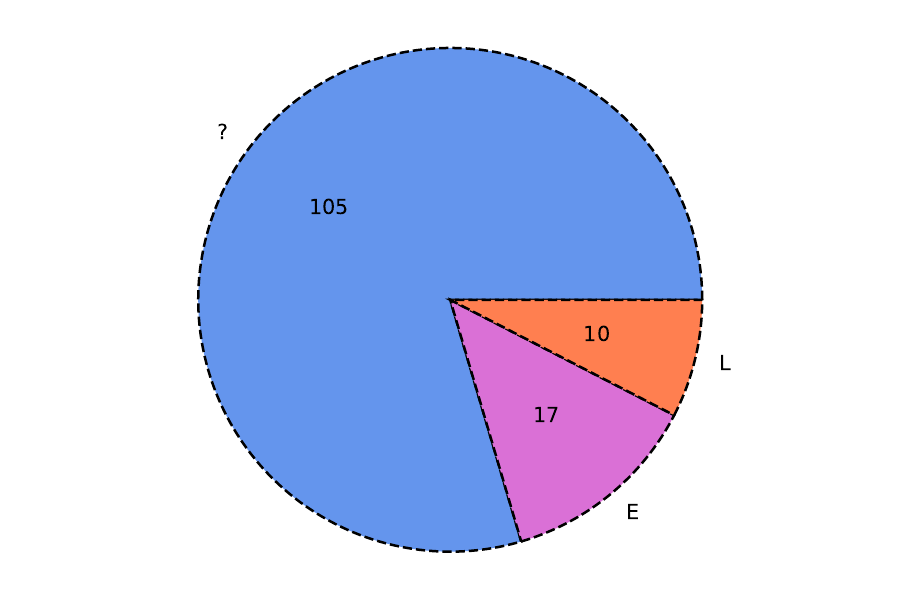}
    \caption{
        Pie chart categorising SKE methods presence/lack of software implementations.
        There, `L' denotes the presence of a reusable library, `E' denotes experiment code, and `?' denotes lack of known technologies.
    }
    \label{fig:pie-ske-tech}
\end{SCfigure}

Among the \skecount{} surveyed methods for SKE, we found runnable software implementations for \skewithtechcount{} (\skewithtechpercent{}\%).
Of these, \skewithlibcount{} consist of reusable software libraries, while the others are just experimental code.
\Cref{fig:pie-ske-tech} summarises this situation.
In \Cref{app:summary-ske}, we provide details about these implementations---there including the algorithm they implement and the link to the repository hosting the source code.

\subsection{Symbolic Knowledge Injection}
\label{subsec:ski-taxonomy}


As far as SKI is concerned, we take into account no prior taxonomy.
Indeed, despite the methods surveyed in this subsection come from well-studied (yet disjoint) research communities -- such as neuro-symbolic computation \citep{BesolddABBDHKLLMvlPPPZ2017} and knowledge graph embedding \citep{WangMWG17} --, we are not aware of any prior work attempting to unify these research areas under the SKI umbrella.


Along this line, we cluster the surveyed SKI methods according to four orthogonal dimensions, namely:
\begin{inlinelist}
    \item the type of SK they can inject,
    \item the strategy they follow to attain injection,
    \item the kind of predictors they can be applied to,
    \item the aim they pursue while performing injection.
\end{inlinelist}
In what follows, we answer research questions \ref{item:rq6}--\ref{item:rq10} by focusing on such dimensions, individually.
Conversely, in the supplementary materials (see \Cref{app:summary-ski}), we overview the \skicount{} methods selected for SKI.

\subsubsection{\ref{item:rq7}: Which kind of SK can be injected into ML predictors?}
\label{subsubsec:RQ7}

Generally speaking, SKI methods support the injection of knowledge expressed by various formalisms---despite each surveyed method focuses on some particular formalism.
Along this line, a key discriminating factor is whether the chosen formalism is machine-interpretable or not---other than human-interpretable.

W.r.t.\ the formalism the input knowledge should adopt to support SKI, we may cluster the surveyed methods into two major groups, namely:
\begin{description}
    \item[logic formul\ae{}] or \textbf{knowledge bases (KB)} (i.e., sets of formul\ae{}) adhering to either FOL or some of its subsets, which are therefore both machine- and human-interpretable.
    Here, admissible sub-categories reflect the kinds of logics described in \Cref{subsubsec:formal-logics}.
    Ordered by decreasing expressiveness, these are:
    \begin{description}
        \item[full first-order logic] formul\ae{} including recursive terms, possibly containing variables, predicates of any arity, and logic connectives of any sorts, possibly expressing definitions;

        \item[Horn logic] (a.k.a.\ \textbf{Prolog}-like) where knowledge bases consist of head--body rules, involving predicates and terms of any sorts;

        \item[Datalog] i.e., Horn clauses without recursive terms (only constant or variable terms allowed);
        
        \item[modal logics] i.e., extensions of some logic above with modal operators (e.g., $\Box$ and $\Diamond$), denoting the \emph{modality} in which statements are true (e.g., \emph{when}, in temporal logic); 

        \item[knowledge graphs] i.e., a particular application of description logics aimed at representing entity--relation graphs;

        \item[propositional logic] where expressions are simply expressions involving boolean variables and logic connectives.
    \end{description}

    \item[expert knowledge] i.e., any piece of human- (but not necessarily machine-) interpretable knowledge by which data generation can be attained.
    This might be the case of physics formulae, syntactical knowledge, or any form of knowledge that is usually held by a set of human experts, and, as such, is only accessible to human beings.
    For this reason, expert knowledge injection requires some data to be generated to reify its information in tensorial form.
    Of course, expert knowledge may be cumbersome to extract and requires human engineers to take care of data generation before any injection can occur.
\end{description}
%

\begin{SCfigure}
    \centering
    \includegraphics[width=.4\linewidth]{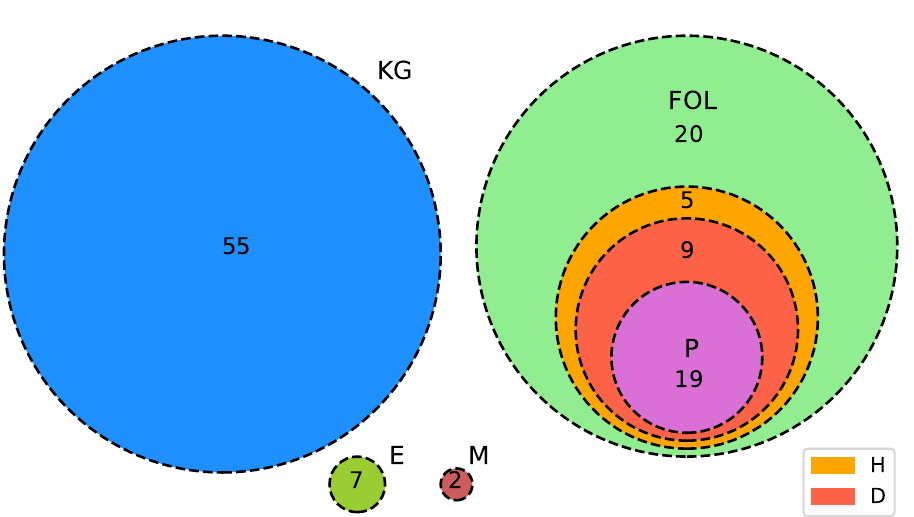}
    \caption{Venn diagram categorising SKI methods w.r.t.\ the \emph{input knowledge} type: knowledge graphs (KG), propositional logic (P), first-order logic (FOL), expert knowledge (E), Datalog (D), Horn logic (H), or modal logic (M).}
    \label{fig:pie-ski-logic}
\end{SCfigure}

In \Cref{fig:pie-ski-logic} we categorise the surveyed SKI methods w.r.t.\ their formalism of choice.
There, KG are the most prominent cluster (including almost a half of the surveyed methods), whereas model logic is the smallest one.
Methods tailored on FOL or its subsets (apart from KG) form another relevant cluster.
Among the FOL subsets, propositional logic plays a pivotal role, as it involves the relative majority of methods.
%

%
Notably, as long as the logic formalism is concerned, we consider and report the \emph{actual} logic used in the papers.
Indeed, this is rarely explicitly stated by the authors into their papers.
So, we deduce the actual logic used by each SKI method from the constraints its logic is subject to, according to its authors.


\subsubsection{\ref{item:rq9}: How does SKI work?}
\label{subsubsec:RQ9}


\begin{figure}
    \centering
    \caption{Overview of major strategies followed by surveyed SKI methods.}
    \label{fig:ski-strategies}

    \subfloat[Structuring strategy: a (portion of a) neural network is constructed, mirroring the symbolic knowledge.]{
        \includegraphics[width=0.5\linewidth]{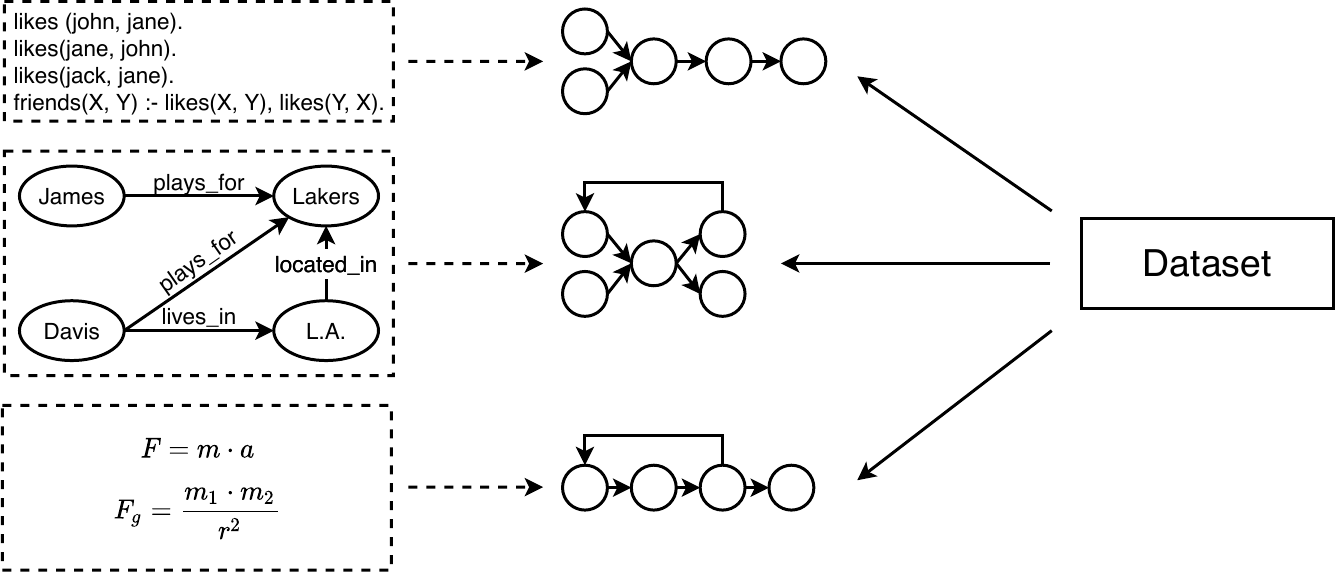}
        \label{fig:ski-structuring}
    }
    \subfloat[Guided learning strategy.]{
        \centering
        \includegraphics[width=0.4\linewidth]{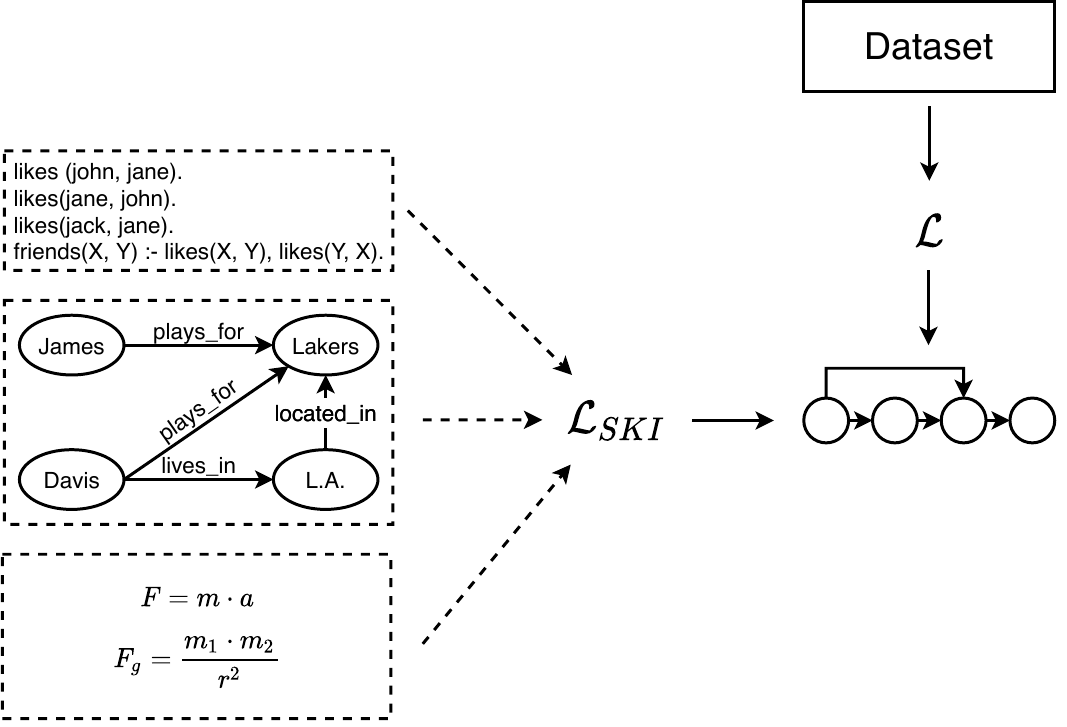}
        \label{fig:ski-loss}
    }
    
    \subfloat[Embedding strategy: the symbolic knowledge is converted in tensorial form and ML predictors are fed ``as usual''.]{
        \includegraphics[width=0.7\linewidth]{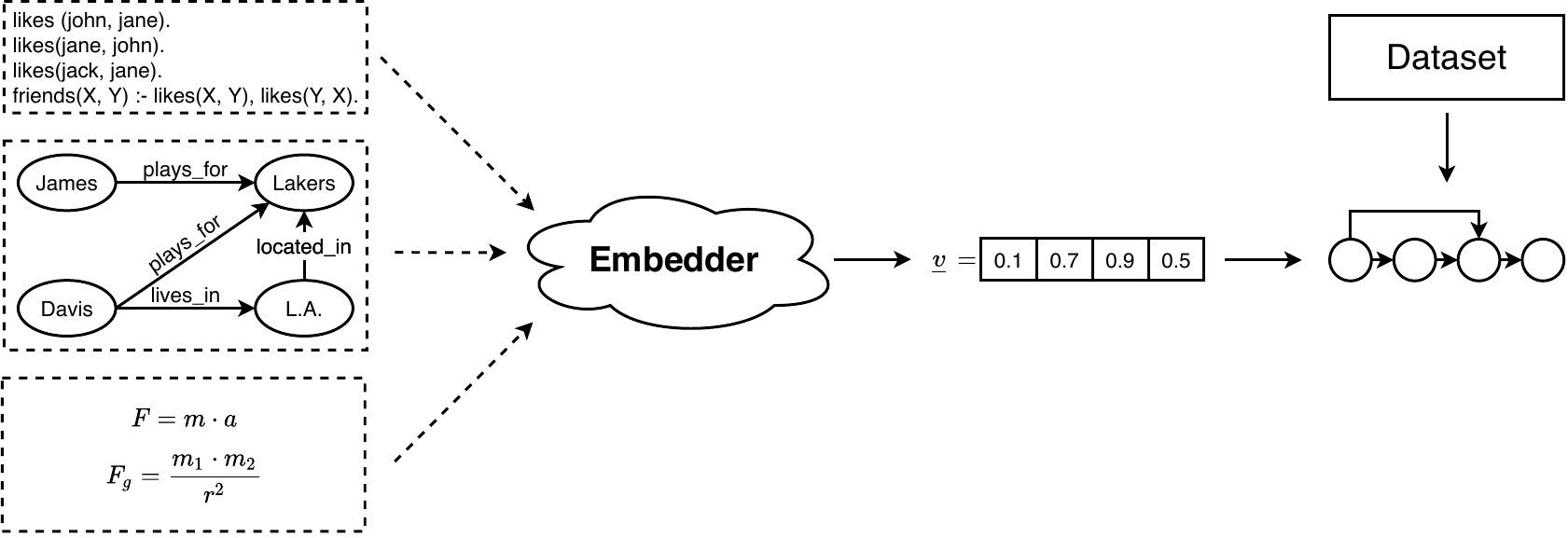}
        \label{fig:ski-embedding}
    }
\end{figure}

\begin{SCfigure}
    \centering
    \includegraphics[width=.3\linewidth]{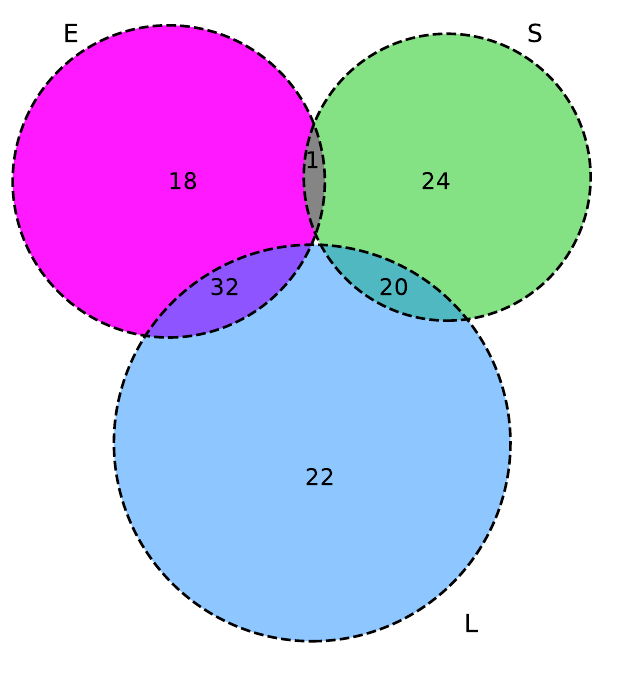}
    \caption{Venn diagram categorising SKI methods w.r.t.\ \emph{strategy}: structuring (S), embedding (E), or guided learning (L).}
    \label{fig:pie-ski-integration}
\end{SCfigure}

By analysing the surveyed SKI methods, we acknowledge great variety in the actual way injection is performed.
Arguably, however, such variety can be tackled by focusing on tree major \emph{strategies}, depicted in \Cref{fig:ski-strategies} and summarised below:
%
%
\begin{description}
    \item[predictor structuring] where (a part of) a sub-symbolic predictor (commonly, NN) is created to mirror the symbolic knowledge via its own internal structure.
    %
    In other words, a predictor is created or extended to mimic the behaviour of the SK to be injected.
    For instance, when it comes to NN, their internal structure is crafted to represent logic predicates via neurons, and logic connectives via synapses;

    \item[knowledge embedding] where SK is converted into numeric-array form -- e.g., vectors, matrices, tensors, etc. -- to be provided as `ordinary' input for the sub-symbolic predictor undergoing injection.
    In other words, numeric data is generated out of symbolic knowledge.
    Any numeric representation of this sort is called \emph{embedding} [of the original symbolic knowledge].
    %
    %
    For example, this is the common strategy exploited by the knowledge graph embedding community \citep{WangMWG17}, as well as by graph NN \citep{gnn-woa2021,LambGGPAV20};

    \item[guided learning] (a.k.a., \textbf{constraining}) where SK is used to steer the learning process of ML predictors, by either penalising inconsistent behaviours or by incentivising consistent behaviours w.r.t.\ the SK.
    When the predictor undergoing injection is trained via some optimisation process involving loss functions being minimised (e.g., NN), guided learning is achieved by altering those loss functions in such a way that violations w.r.t.\ the SK increase the loss.
    A dual statement holds for predictors requiring training to step through maximization processes.
    %
    %
    %
    %
    %
    %
    The recent book by \cite{Gori2018} nicely overviews methods of these kinds.

\end{description}
\Cref{fig:pie-ski-integration} summarises the frequency of these strategies among the surveyed SKI algorithms.
Notably, the distribution of surveyed SKI methods among the three categories above is quite balanced. 


\subsubsection{\ref{item:rq6}: Which sorts of ML predictors can SKI be applied to?}
\label{subsubsec:RQ6}

Virtually \emph{all} surveyed SKI methods are designed to inject knowledge into NN.
However, as this survey spans over 2 decades, the sorts of NN supported by SKI methods are manifold---despite each method is tailored on specific sorts of NN.

\begin{SCfigure}
    \centering
    \includegraphics[width=.4\linewidth]{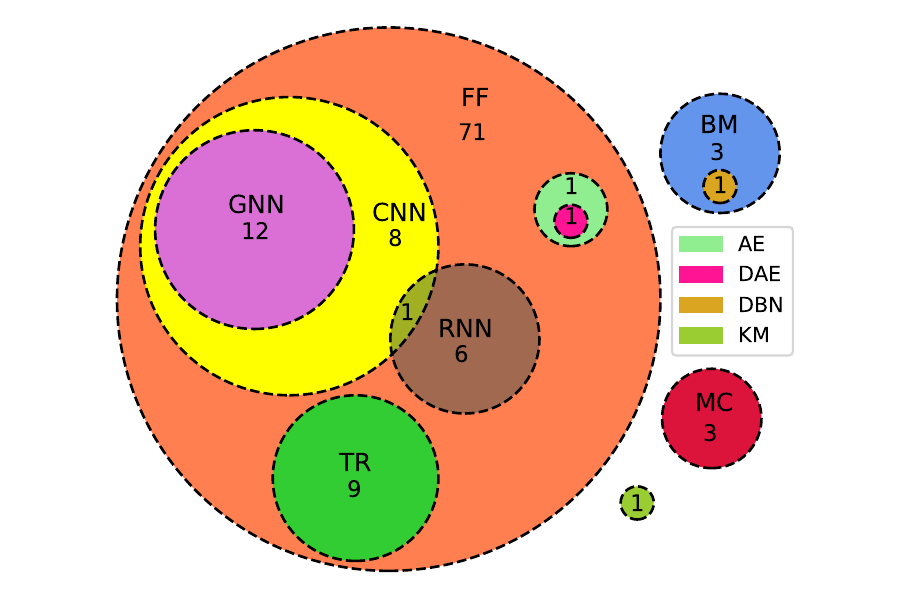}
    \caption{Venn diagram categorising SKI methods w.r.t.\ the \emph{targeted predictor} type: feed-forward (FF), convolutional (CNN), graph (GNN) or recurrent (RNN) neural networks, Boltzmann machines (BM), Markov chains (MC), transformers (TR), auto-encoders (AE), deep belief networks (DBN), denoising auto-encoders (DAE), kernel machines (KM).}
    \label{fig:pie-ski-predictors}
\end{SCfigure}

Accordingly, surveyed SKI methods can be classified w.r.t.\ the particular sort of NN they support.
As detailed by \Cref{fig:pie-ski-predictors}, admissible choices along this line fit the many sorts of NN discussed in \Cref{par:nn}, namely:
\begin{description}
    \item[feed-forward NN] multi-layered NN where neurons from layer $i$ are only connected with layer $i+1$, and multiple ($\geq 2$) layers may exist;
    \item[convolutional NN] particular cases of feed-forward NN, involving convolutional layers as well;
    \item[graph NN] particular cases of convolutional NN tailored on graph-like data;
    \item[recurrent NN] particular cases of NN admitting loops among layers;
    \item[Boltzmann machine] a particular neural architecture where connections are undirected---i.e., every node is connected to every other node;
    \item[transformer] particular case of NN that leverage a self-attention mechanism---i.e., differentially weighting parts of the input data depending on their significance;
    \item[auto-encoders] particular case of feed-forward NN, characterised by a bottleneck architecture used to learn reduced data encodings through learning to regenerate the input from the encoding;
    \item[deep belief networks] a composition of multiple Boltzmann machines, stacked altogether, in a feed-forward fashion;
    \item[denoising auto-encoder] particular case of auto-encoders working over corrupted input.
\end{description}
Notable exceptions are:
\begin{description}
    \item[kernel machines] ML models relying on kernels---i.e., similarity measures between observed patterns;
    \item[Markov chains] state machines with probabilities on state transitions, modelling stochastic phenomena.
\end{description}
%

The reason why the vast majority of methods rely on (some sort of) NN is straightforward: methods tailored upon GNN (resp.\ CNN) assume the networks to accept specific kinds of data as input, e.g.\ graphs (resp.\ images), while ordinary feed-forward NN accept raw vectors of real numbers.

\subsubsection{\ref{item:rq8}: For which kind of AI tasks can SKI be exploited?}
\label{subsubsec:RQ8}


\begin{SCfigure}
    \centering
    \includegraphics[width=.4\linewidth]{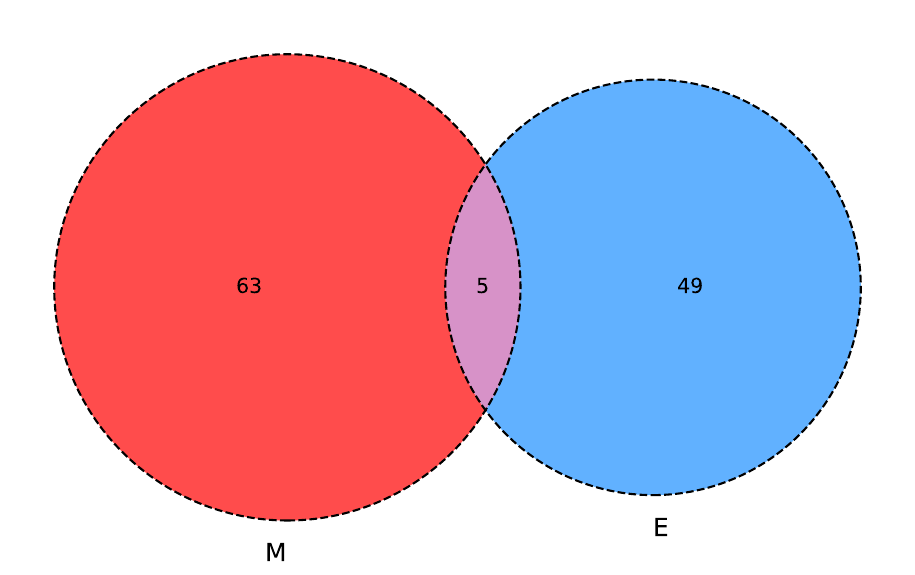}
    \caption{Venn diagram categorising SKI methods w.r.t.\ \emph{aim}: knowledge manipulation (M) or enrich (E).}
    \label{fig:pie-ski-aim}
\end{SCfigure}

Unlike SKE methods -- which uniquely serve the purpose of inspecting black-box predictors by mimicking the way they address supervised learning tasks --, SKI methods from the literature may serve multiple purposes. 
As outlined by \Cref{fig:pie-ski-aim}, we identify two major purposes SKI methods may pursue, by targeting either symbolic or sub-symbolic AI tasks.
More precisely, SKI methods may pursue:
\begin{description}
    \item[symbolic knowledge manipulation] where SKI enables the \emph{sub}-symbolic manipulation of \emph{symbolic} knowledge, by letting sub-symbolic predictors treat SK similarly to what done by symbolic engines.
    In doing so, SKI supports symbolic-AI tasks such as
    \begin{description}
        \item[logic inference] in its many forms (e.g.\ deductive, inductive, probabilistic, etc.), i.e.\ drawing conclusions out of symbolic KB;
        \item[information retrieval] looking for information in symbolic KB;
        \item[KB completion] finding (and adding) missing information in symbolic KB;
        \item[KB fusion] merging several KB into a single one, taking care of (possibly, syntactically different) overlaps;
    \end{description}
    The key point here is supporting tasks where both inputs and outputs are symbolic in nature, but leveraging upon sub-symbolic methods to gain speed, fuzziness, and robustness against noise.
    \item[learning support] (a.k.a., \textbf{enrich}) where SKI lets \emph{sub}-symbolic methods consume \emph{symbolic} knowledge to either improve or enrich learning capabilities.
    In doing so, SKI supports ordinary ML tasks -- such as classification --, by allowing ML predictors to process
    (or take advantage by) structured symbolic knowledge.
    The underlying idea of such approaches is that there exist some concepts that are cumbersome or troublesome to learn from examples---e.g., syntactical concepts, semantics, etc.
    Therefore, SK expressing these high-level concepts may be injected directly into the model to be trained.
\end{description}
As the reader may note from the picture, surveyed SKI methods are quite balanced w.r.t.\ the categories above, with a slight preference for SK manipulation.

\subsubsection{\ref{item:rq10}: which and how many SKI algorithms come with runnable software implementations?}


\begin{SCfigure}
    \centering
	\includegraphics[width=.35\linewidth]{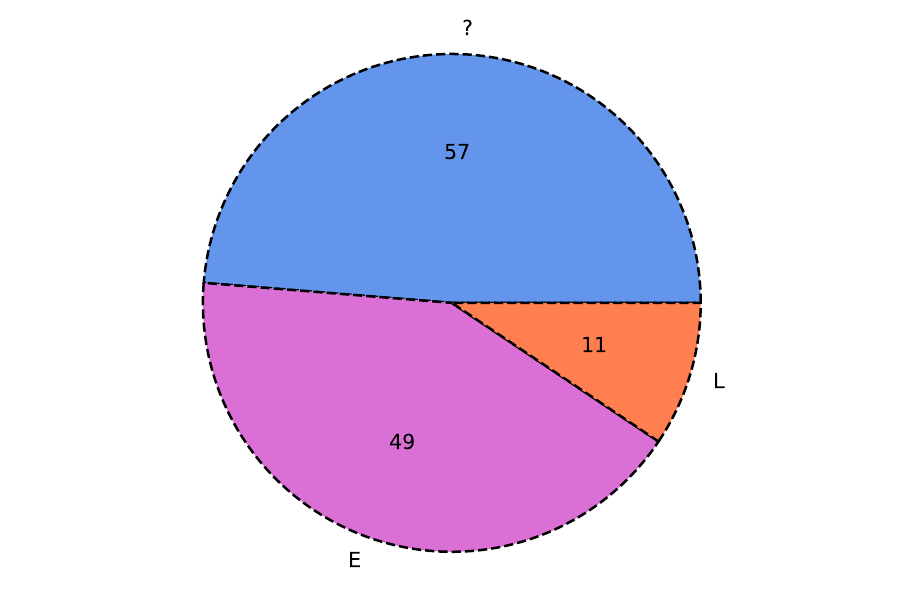}
    \caption{
        Pie chart categorising SKI methods presence/lack of software implementations.
        There, `L' denotes the presence of a reusable library, `E' denotes experiment code, and `?' denotes lack of known technologies.
    }
    \label{fig:pie-ski-tech}
\end{SCfigure}

Among the \skicount{} surveyed methods for SKI, we found runnable software implementations for \skiwithtechcount{} (\skiwithtechpercent{}\%).
Of these, \skiwithlibcount{} consist of reusable software libraries, while the others are just experimental code.
\Cref{fig:pie-ski-tech} summarises this situation.
In the supplementary materials (see \Cref{app:summary-ski}), we provide details about these implementations---there including the algorithm they implement and the link to the repository hosting the source code.

\section{Discussion}
\label{sec:discussion}

\begin{figure}
    \includegraphics[width=\linewidth]{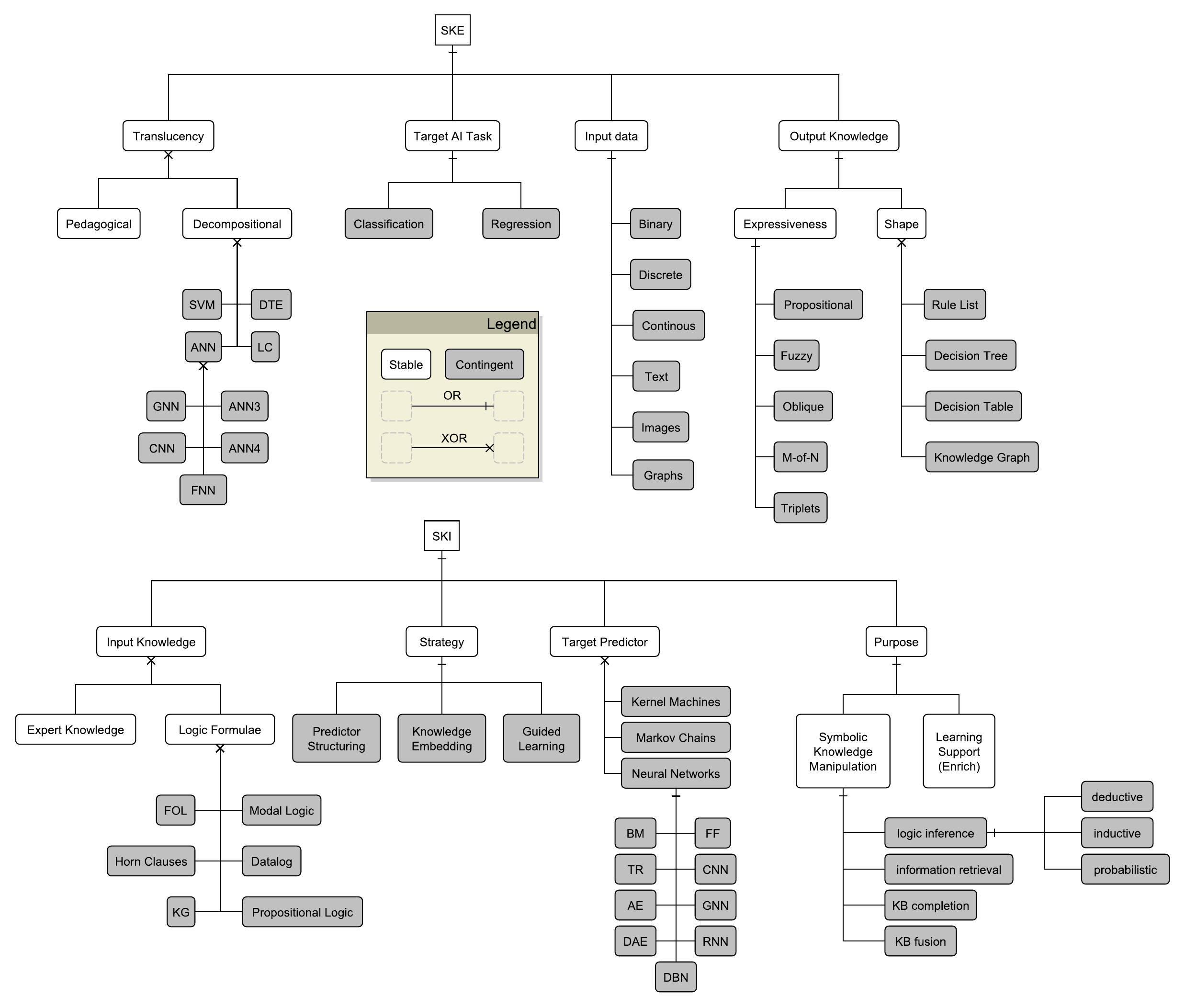}
    \caption{Summary of SKE and SKI taxonomies derived from the literature, as discussed in \cref{sec:results}.}
    \label{fig:taxonomies}
    \label{fig:ske-taxonomy}
    \label{fig:ski-taxonomy}
\end{figure}

\Cref{fig:taxonomies} summarises the main contribution of our paper---i.e., the taxonomies for SKE and SKI we induced from the surveyed literature.
Generally speaking, such taxonomies are useful tools to categorise present (and, hopefully, future) SKE/SKI methods, and to highlight the relevant features of each particular method.
In this way, the interested readers may figure out what to expect from any given SKE/SKI method, as well as draw general analyses concerning the state of the art.
Accordingly, in this section we analyse our taxonomies, elaborating on the current challenges and future perspectives.

It is worth mentioning that our taxonomies involve both `stable' and `contingent' categories by which SKE/SKI methods can be described.
These are represented as either white or grey boxes in \Cref{fig:taxonomies}.
Stable categories are time-independent and they are not susceptible to change in the near future, while contingent categories are subject to trends and may evolve.
Consider for instance SKE methods (see \Cref{fig:ske-taxonomy}), categorised w.r.t.\ their output knowledge.
While expressiveness is a stable sub-category, its actual sub-sub-categories are contingent, meaning that new ones may be added in the future.

\subsection{SKE Taxonomy}

As shown in \Cref{fig:ske-taxonomy}, SKE methods can be classified by
\begin{inlinelist}
    \item\label{item:ske-translucency} translucency,
    \item\label{item:ske-task} targeted AI task,
    \item\label{item:ske-input} nature of the input data, and
    \item\label{item:ske-output} form of the output knowledge.
\end{inlinelist}
W.r.t.\ \cref{item:ske-translucency}, SKE methods can either be categorised as pedagogical or decompositional.
In the particular case of decompositional methods, the actually targeted predictor is relevant too---and possibilities \emph{currently} include NN, DT, SVM, and linear classifiers.
W.r.t.\ \cref{item:ske-task}, SKE methods may target classification or regression tasks, or both.
In any case, they \emph{currently} target supervised ML tasks alone.
W.r.t.\ \cref{item:ske-input}, SKE methods accept predictors trained upon binary, discrete, or continuous data, as well as images, graphs and text.
Finally, w.r.t.\ \cref{item:ske-output}, SKE methods may produce symbolic knowledge of different shapes, and with different expressiveness.
Shapes may \emph{currently} involve rule lists, as well as graphs, decision trees or tables.
Conversely, as long as expressiveness is involved, symbolic knowledge may be propositional or fuzzy -- possibly including \mofn{}-like statements -- , or be expressed as triplets or oblique rules.

\paragraph{About translucency}

It is worth stressing the relevance of pedagogical methods from the engineering perspective.
Indeed, if properly implemented, pedagogical methods may be exploited in combination with predictors of any sorts.
Of course, they are expected to reach lower performances w.r.t.\ decompositional ones, as they access less information.
On the other side, decompositional methods may be more precise at the expense of generality.

\paragraph{About input data}

We recall that binary features are particular cases of discrete features, while discrete features are, in turn, particular cases of continuous features.
Hence, it is worthwhile noticing that extractors requiring only binary features can be applied to categorical datasets by pre-processing discrete attributes via one-hot encoding (OHE).
Analogously, extractors requiring discrete features can work with continuous attributes if those continuous features are \emph{discretised}.
Finally, continuous features can be converted into binary ones by performing discretisation and OHE, in this exact order.

While these transformations can always be applied in the general case, some authors have included them in their SKE methods at the design level.
Hence, some papers explicitly account discretisation or OHE as part of the SKE methods they propose.
This is the case, for instance, of the methods enclosed in the intersection between the `C' and `D' sets in \Cref{fig:pie-ske-input} (and labelled as `C+D' in the supplementary materials, \Cref{app:supplementary}).
Other methods may instead rely upon other discretisation strategies, such as the ones surveyed by \cite{YangWW10}.

\paragraph{About output knowledge}

It is worth stressing that differences among rule lists, decision trees, and tables are mostly syntactic, as conversions among these forms are possible in the general case (cf. \Cref{app:examples} for examples).
%
As far as expressiveness is concerned, we remark that all logic formalisms currently in use for SKE are essentially particular cases of propositional logic---possibly, under a fuzzy interpretation.
This implies that the full power of FOL is far from being fully exploited in practice.

Finally, we point out some correlations among the expressiveness of output rules and the nature of the predictor they are extracted from, as well as the input data it is trained upon.
For instance, SKE methods working with \emph{continuous} input data are more likely to adopt oblique rules---or, at least, propositional rules with arithmetic comparisons.
In fact, decisions are there drawn by comparing numeric variables with constants or among each other.
Another example: some \emph{decompositional} SKE methods focusing upon NN adopt \mofn{} statements.
Arguably, the reason is that \mofn{} expressions aggregate several elementary statements into a single formula, similarly to how neurons aggregate synapses from previous layers in NN---hence such methods approximate neurons via \mofn{} expressions.

\paragraph{On SKE methods' chronology}

In conclusion, we stress the chronological distribution of SKE methods.
As highlighted in \Cref{app:summary-ske}, the majority of SKE methods have been proposed during the decades ranging from the 90s up to the 2010s.
Contributions slowed down after that, up to the 2020s, where SKE gained new momentum.

In our opinion, research on ML interpretability gained momentum more than once in the history of AI.
Each time sub-symbolic AI attracted the interest of researchers, so did the need to make it more comprehensible.
Arguably, this is the reason why most SKE-related works are concentrated around the 2000s. 
In the last years, we are witnessing the novel spring of sub-symbolic AI \citep{Maclure20}, which is, in turn, motivating researchers' interest in XAI.
Arguably, this is why SKE is gaining novel momentum in recent years.

\subsection{SKI Taxonomy}

As shown in \Cref{fig:ski-taxonomy}, SKI methods can be classified by
\begin{inlinelist}
    \item\label{item:ski-input} form of the input knowledge,
    \item\label{item:ski-strategy} followed strategy,
    \item\label{item:ski-target} targeted predictor type, and
    \item\label{item:ski-purpose} purpose.
\end{inlinelist}
W.r.t.\ \cref{item:ski-input}, SKI methods can either accept logic formul\ae{} or expert knowledge as input.
In the former case, \emph{current} possibilities include FOL and its subsets, and in particular knowledge graphs.
W.r.t.\ \cref{item:ski-strategy}, SKI methods may \emph{currently} follow one of three strategies, namely: predictor structuring, knowledge embedding, or guided learning.
W.r.t.\ \cref{item:ski-target}, SKI methods \emph{currently} mostly target NN-based predictors, other than Markov chains and kernel machines.
Finally, w.r.t.\ \cref{item:ski-purpose}, SKI methods may pursue two kinds of purposes, non-exclusively: manipulating symbolic knowledge or supporting/enriching learning.
In the former case, \emph{current} possibilities involve symbolic AI related tasks such as logic inference (and its many forms), information retrieval, and KB completion/fusion.

\paragraph{About input knowledge and injection strategies}

Logic formul\ae{} are the most common approach to define prior concepts to be injected.
This is true, in particular, for SKI approaches following the model structuring or guided learning strategies.
Indeed, via logic formul\ae{}, they express criteria that sub-symbolic models should satisfy or emulate.
However, methods of these sorts often require formul\ae{} to be grounded.
Grounding introduces computational burden and hinders capability of representing recursive or infinite data structures---hence limiting what can actually be injected.

Conversely, knowledge graphs are the most common knowledge representation approach when it comes to perform SKI following the knowledge embedding strategy.
This is unsurprising, given that `knowledge graph embedding' is a research line \emph{per se}.
%
%

\paragraph{About target predictors}

Neural networks play a pivotal role in SKI.
%
%
Arguably, the reason lies in the great \emph{malleability} of NN w.r.t.\ their structure and training, as well as their \emph{flexibility} w.r.t.\ feature learning.
In fact, NN come in different shapes as different architectures may be constructed by connecting neurons in various ways.
This is fundamental to support SKI via \emph{predictor structuring}.
Furthermore, as long as their architectures are DAG, NN can be trained via gradient descent, i.e.\ by minimising a loss function of arbitrarily defined.
This is, in turn, fundamental to support SKI via \emph{guided learning}.
Finally, feature learning is a characterising capability of NN, making them capable to automatically elicit the relevant aspects they should focus upon, w.r.t.\ input data.
This is the reason why NN are well suited for the knowledge embedding strategy as well.
Accordingly, to the best of our knowledge, there exists no other sort of predictor having similar flexibility and malleability.

\paragraph{On SKI methods' chronology}

In conclusion, we stress the chronological distribution of SKI methods.
As highlighted in \Cref{app:summary-ski}, the majority of SKI methods have been proposed after 2010, and, notably, the amount of contribution has exploded after 2015.


In our opinion, this distribution is due to the composite effect of three major drivers, namely: natural language processing (NLP), XAI, and neuro-symbolic computation (NSC).
Arguably, all such drivers are gaining momentum in the last years, due to the success of machine- and deep-learning.
Indeed, NLP reached unprecedented performance levels after it started leveraging on DL, possibly combined with knowledge graphs and the corresponding SKI methods.
Similarly, a portion of XAI-related contributions propose SKI methods aimed at controlling, constraining, or guiding what predictors learn from data.
Finally, NSC has recently emerged as a field exploiting SKI methods to process logic knowledge sub-symbolically, by exploiting the malleability of NN.

\subsection{Challenges}

We observe that SKE algorithms focus exclusively on supervised learning tasks -- i.e., classification and regression -- while they do not tackle unsupervised or reinforcement learning tasks---e.g., clustering or optimal policy search.
One may argue, for instance, that clustering algorithms are not opaque -- e.g., $K$-nearest neighbours --, despite operating on numeric data.
However, \emph{pedagogical} SKE algorithms could be used on clustering predictors with no or minimal adjustments---as trained clustering predictors are essentially classifiers upon anonymous classes.
Similarly, it could be possible to perform extraction on predictors trained using reinforcement learning with existing SKE algorithms.
Indeed, future literature works on SKE for unsupervised learning would be needed.

Furthermore, the vast majority of SKI algorithms accept knowledge in form of knowledge graph -- a.k.a., description logic -- or propositional logic (\Cref{fig:pie-ski-logic}), which are much less expressive than FOL.
These logics lack support for recursion and function symbols, meaning that the user is quite limited in providing knowledge to predictors.
The reason is that common ML predictors are acyclic (e.g., NN, etc.), meaning that there is no straightforward way to integrate recursion nor indefinitely deep data structures without severe information loss due to approximations.
Hence, future research efforts concerning SKI may consider addressing the injection of logics involving recursive clauses or arbitrarily deep data structures.

\subsection{Opportunities}
\label{ssec:opportunities}

We propose a brief discussion on the benefits arising from the \emph{joint} exploitation of both SKI and SKE in the engineering of AI solutions.
%
\begin{inlinelist}
    \item the possibility of \emph{debugging} sub-symbolic predictors, and
    \item the exploitation of symbolic knowledge as the \emph{lingua franca} among heterogeneous hybrid systems.
\end{inlinelist}
Accordingly, in the remainder of this sub-section, we delve into the details of these expected benefits.

\subsubsection{Debugging sub-symbolic predictors}

Debugging is a common activity for software programmers: it aims at spotting and fixing \emph{bugs} in computer programs under production/maintenance.
There, a bug is some unknown error contained into the program which leads to some unexpected or undesired observable behaviour of the computer(s) running that program.
The whole procedure relies on the underlying assumption that computer programs are intelligible to the programmer debugging them, and that the program can be precisely edited to fix the bug.

One may consider XAI techniques as means to `debug' sub-symbolic predictors.
In this metaphor, sub-symbolic predictors correspond to computer programs -- despite they are not manually written by programmers, but learned from data --, while data scientists correspond to programmers.
However, debugging sub-symbolic predictors is hard, because of their opacity -- which makes their inner behaviour poorly intelligible for data scientists --, and because they cannot be precisely edited after training---but should be rather retrained from scratch.
Accordingly, we discuss here the role of SKE and SKI in overcoming these issues, hence allowing data scientists to debug sub-symbolic predictors.

\begin{figure}
    \includegraphics[width=.8\linewidth]{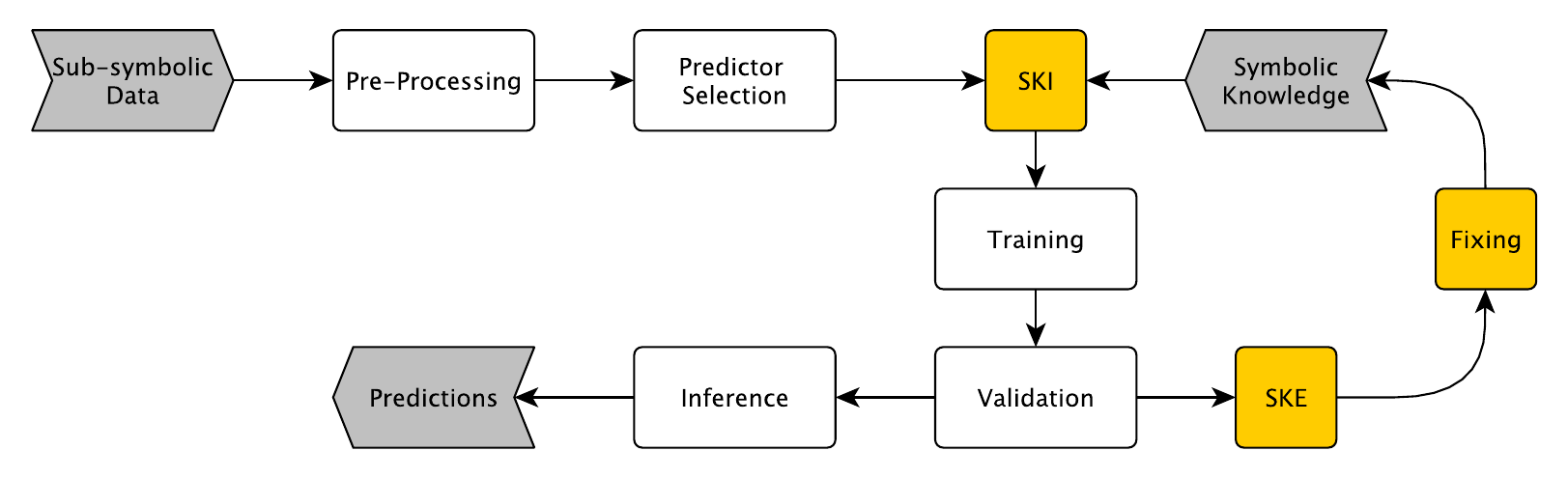}
    \caption{ML workflow enriched with SKI and SKE phases. On the right, the train-extract-fix-inject loop is represented.}
    \label{fig:ml-workflow-ske-ski}
\end{figure}

\Cref{fig:ml-workflow-ske-ski} provides an overview of how SKI and SKE fit the generic ML workflow.
In particular, the figure stresses the relative position of both SKI and SKE w.r.t.\ the other phases of the ML workflow.
Notably, SKI should occur before (or during) training, while SKE should occur after it.
However, the figure also stresses the addition of a \emph{loop} into an otherwise linear workflow (right-hand side of the figure).
We call it the `train-extract-fix-inject' (TEFI) loop, and we argue it is a possible way to debug sub-symbolic predictors.

In the TEFI loop, SKE is the basic mechanism by which the inner operation of a sub-symbolic predictor (i.e., `the program', in the metaphor) is made intelligible to data scientists.
The extracted knowledge may then be understood by data scientists and `debugged'---hence looking for pieces of knowledge which are wrong w.r.t.\ the data scientist expectations.
Then, symbolic knowledge may be precisely edited and fixed.
Along this line, SKI is the basic mechanism by which a trained predictor is precisely edited to adhere to the fixed symbolic knowledge.

\subsubsection{Symbolic knowledge as the \emph{lingua franca} for intelligent systems}


Intelligent systems can be suitably modelled and described as composed of several intelligent, heterogeneous, and hybrid computational agents interoperating -- and possibly communicating -- among each others.
There, a computational agent is any software or robotic entity capable of computing, other than perceiving and affecting some given environment---be it the Web, the physical world, or anything in between.
To make the overall systems intelligent, these agents should be capable of a number of intelligent behaviours, ranging from image, speech, or text recognition to autonomous decision making, planning, or deliberation.
Behind the scenes, these agents may (also) leverage upon sub-symbolic predictors -- possibly trained upon locally-available data --, as well as symbolic reasoners, solvers, or planners to support such sorts of intelligent behaviours.
In this sense, such agents are \emph{hybrid}, meaning that they involve both symbolic and sub-symbolic AI facilities.
However, interoperability may easily become a mirage because of
\begin{inlinelist}
    \item the wide variety of algorithms, libraries, and platforms supporting sub-symbolic ML, other than
    \item the possibly different data items each agent may locally collect and later train predictors upon.
\end{inlinelist}
Indeed, each agent may learn (slightly) different behaviours, due to the differences in the training data and in the actual ML workflow it locally adopts.
When this is the case, exchange of behavioural knowledge may become cumbersome or infeasible.

In such scenarios, SKI and SKE may be enablers of a higher degree of interoperability, by supporting the exploitation of symbolic knowledge as the \emph{lingua franca} for heterogeneous agents.
Indeed, hybrid agents may exploit SKE to extract symbolic knowledge out of their local sub-symbolic predictors, and exchange (and possibly improve) that symbolic knowledge with other agents.
Then, any possible improvement of the symbolic knowledge attained via interaction may be back-propagated into local sub-symbolic predictors via SKI, hence enabling agents' behaviour to improve as well.

\subsection{Limitations}

This SLR also means to be as comprehensive, precise, and reproducible as possible; nonetheless, we acknowledge two potential limitations: \emph{(i)} the expected life span of our taxonomies, and \emph{(ii)} terminology issues in the literature.

Both SKE and SKI are becoming increasingly popular topics nowadays: further advancements have to be expected for the next decade, at least.
Hence, our taxonomies may require to be verified, and possibly updated, sometime in the future.
The straightforward methodological approach defined by our SLR, however, should ensure a  clear path to future reproductions of this work.

Also,  an evolution in the naming conventions clearly emerge from our analysis.
Along the years, SKE has been called in disparate ways---e.g., ``rule extraction'' \cite{andrews1995survey}, or ``knowledge distillation'' \cite{cheng2021}, just to name a few.
The same holds for SKI: there, naming conventions are commonly based on the injection strategy, yet they rarely contain the word `injection'.
So, we may have missed some works while collecting papers simply because they were using different naming conventions that we were not able to devise out.
This is an inherent issue of the keyword-based methodology we adopted for SLR.
To minimise issues in the classifications of present and future SKE/SKI methods, we draw loose definitions, and carefully read papers to determine whether they match our definitions or not.
Yet, the existence of missing works for unexpected terminology choices cannot be excluded.

\section{Conclusion}
\label{sec:conclusion}

%


In this paper we survey the state of the art of symbolic knowledge extraction and injection under a XAI perspective.
Stemming from two original definitions, we \emph{systematically} explore the literature of both SKE and SKI, spanning a period of four decades.
Our goal is to elicit the major characteristics of SKE/SKI algorithms form the literature (\ref{item:SKE}--\ref{item:SKI}), hence deriving general taxonomies which we hope other researchers may exploit.
Another goal of ours is to assess the current state of technologies (\ref{item:tech}) by identifying software implementations of SKE/SKI techniques.

Considerable efforts were spent in keeping our review \emph{reproducible}---as prescribed by the goal question metric approach by \cite{goal-question-metric}.
Along this line, we design ten research questions (\ref{item:rq1}--\ref{item:rq10}), and we engineer \emph{ad hoc} queries to be performed on most relevant search engines for scientific literature.
We select \allcount{} primary works, almost evenly distributed among SKE and SKI, other than 11 secondary works.
By analysing these papers, we define and discuss two general taxonomies for both SKE and SKI, which are general enough to categorise present (and possibly future) methods.

Roughly, surveyed methods are categorised w.r.t.\ what they accept as input and produce as output (in terms of symbolic knowledge or predictors), other than how they operate and why.
We also collect data about which and how many SKE/SKI methods come with runnable software implementations (namely, \allwithtechcount{}, i.e., \allwithtechpercent{}\%).
In \Cref{app:supplementary}, we also report Web homepages for the available implementations.

Overall, the implications of our study are manifold.
It demonstrates how SKE and SKI are hot topics of AI research, nowadays.
Literature already contains hundreds of contributions, and our taxonomies provide an effective tool for navigating it.
Hopefully, our SLR can also serve as a map for future contributions, which we expect to flourish soon and abundantly. 
Along this line, our survey summarises what has already been done, and what is currently lacking (cf.\ \cref{ssec:opportunities}).

\begin{acks}
This paper was partially supported by
\begin{inlinelist}
    \item the CHIST-ERA IV project ``EXPECTATION'' (CHIST\-ERA-19-XAI-005), funded by the Italian MUR;
    \item the ``FAIR—Future Artificial Intelligence Research'' project (PNRR, M4C2, Investimento 1.3, Partenariato Esteso PE00000013), funded by the European Commission under the NextGenerationEU programme;
    \item the “ENGINES — ENGineering INtElligent Systems around intelligent agent technologies” project funded by the Italian MUR program ``PRIN 2022'' under grant number 20229ZXBZM.
\end{inlinelist}
\end{acks}

\renewcommand{\thefigure}{\Alph{figure}}
\renewcommand{\thetable}{\Alph{table}}

\newcommand{\techlegend}{Column ``Tech.'' reports the availability/lack of some software technology implementing the algorithm. 
There, `\libavailable' denotes the presence of a reusable sofware library, whereas `\available' denotes the presence of experimental code, and `\unavailable' denotes lack of known technologies.}

\appendix

\section{Supplementary Material}
\label{app:supplementary}

\subsection{Decision Tables, Trees, and Rule Lists}
\label{app:examples}

\begin{wrapfigure}{r}{.4\linewidth}
	\centering
        \includegraphics[width=.56\linewidth]{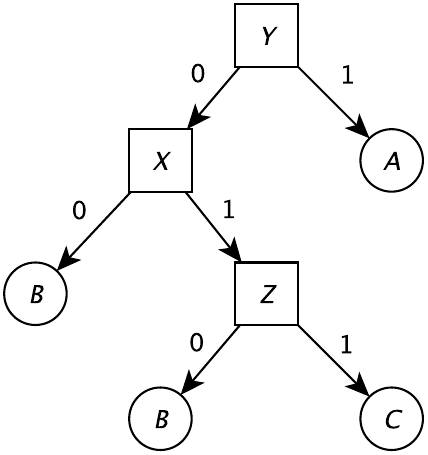}
	\caption{Decision tree equivalent to the decision table reported in \Cref{tab:table-example}.}
	\label{fig:tree-conversion}
\end{wrapfigure}

Decision tables are concise visual representations of rules, specifying one or more conclusions for each set of different conditions. 
They have a row for each input and output variable, and a column for each rule.
Each cell $c_{ij}$ contains the value of the $i$-th variable for the $j$-th rule.
An example of decision table is provided in \Cref{tab:table-example}.
%

\begin{table}
    \caption{
        Example of decision table with 3 binary input variables and 1 output feature---namely, the class label.
        ``Rule set \#1'' and ``Rule set \#2'' are semantically equivalent, but the latter is more compact since it exploits the \emph{don't care} symbol (--).
    }
    \label{tab:table-example}  
    \centering
    \begin{tabular}{c||c|c|c|c|c|c|c|c||c|c|c|c}
        \textbf{Variables} & \multicolumn{8}{c||}{\textbf{Rule set \#1}} & \multicolumn{4}{c}{\textbf{Rule set \#2}}
        \\\hline\hline
        $X$ & 0 & 0 & 0 & 0 & 1 & 1 & 1 & 1 & -- & 0 & 1 & 1 
        \\
        $Y$ & 0 & 0 & 1 & 1 & 0 & 0 & 1 & 1 & 1 & 0 & 0 & 0 
        \\
        $Z$ & 0 & 1 & 0 & 1 & 0 & 1 & 0 & 1 & -- & -- & 0 & 1 
        \\\hline
        Output & $B$ & $B$ & $A$ & $A$ & $B$ & $C$ & $A$ & $A$ & $A$ & $B$ & $B$ & $C$ \\
    \end{tabular}  
\end{table}

Notably, rule lists, decision trees, and decision tables are essentially equivalent in their theoretical expressiveness.
Differences mostly lay in their syntax, i.e., how they represent rules.
Conversions amongst these forms are possible, in the general case.
For instance, the decision table from \Cref{tab:table-example} can be converted into the rule tree depicted in \Cref{fig:tree-conversion}, as well as into the following rule list: 
\[
    \text{if}~ Y = 1 ~\text{then}~ A ~ \text{else if}~ X = 1 \wedge Z = 1 ~\text{then}~ C ~ \text{else}~B.
\]

\subsection{Summary about SKE}
\label{app:summary-ske}

\Cref{tab:ske-taxonomy} summarises our analysis regarding the \skecount{} surveyed SKE methods.
Notably, the table enumerates SKE methods in chronological order (w.r.t.\ publication year), grouping them by five-year periods.
Furthermore, coherently w.r.t.\ the sections above, the table reports the translucency (see \Cref{subsubsec:RQ1}), the supported task (see \Cref{subsubsec:RQ4}), the required input type (see \Cref{subsubsec:RQ2}), and the output format and shape of each surveyed method (see \Cref{subsubsec:RQ3}).
The availability/lack of a public software implementation of each SKE algorithm is also reported.


\newcommand{\myhead}{
	\textbf{\#} & \textbf{Method} & \textbf{Year} & \makecell{\textbf{Trans.}\\(\ref{item:rq1},\ref{item:rq5})} & \makecell{\textbf{Task}\\(\ref{item:rq4})} & \makecell{\textbf{Input}\\(\ref{item:rq2})} & \makecell{\textbf{Express.}\\(\ref{item:rq3})} & \makecell{\textbf{Shape}\\(\ref{item:rq3})} & \makecell{\textbf{Tech.}\\(\ref{item:rq10})}
    \\\hline\hline
}

\newcounter{SkeMethod}
\setcounter{SkeMethod}{1}
\newcommand{\newSkeMethodIndex}{\theSkeMethod\stepcounter{SkeMethod}}

\begin{longtable}{c|p{3cm}|c|c|c|c|c|c|c}
    \caption{
        Summary of the knowledge-extraction algorithms.
        %
        Values from the columns ``Translucency'', ``Task'', ``Input'', ``Expressiveness'', ``Shape'', and ``Technology'' refer the corresponding figures form Sec. 4.1.
        %
    }
    \label{tab:ske-taxonomy}\\
    \myhead
    \endfirsthead
    \caption[]{Summary of the knowledge-extraction algorithms (Continued).}\\
    \myhead
    \endhead
    \newSkeMethodIndex & \citesp{breiman1984classification} & P & C+R & C+D & P & DT & \libavailable\footnote{\label{foot:tree}\url{https://scikit-learn.org/stable/modules/tree.html}}$^,$\footnote{\label{foot:psyke}\url{https://github.com/psykei/psyke-python}}
    \\\hdashline
    \newSkeMethodIndex & \citesp{Quinlan86ID3} & P & C & D & P & DT & \available$^{\ref{foot:tree}}$
    \\\hdashline
    \newSkeMethodIndex & \citesp{SaitoN88} & P & C & D & P & L & \unavailable
    \\\hdashline
    \newSkeMethodIndex & \citesp{ClarkN89} & P & C & C+D & P & L & \available\footnote{\url{https://github.com/alessiamondolo/cn2-rule-based-classifier}}
    \\\hline
    \newSkeMethodIndex & \citesp{Masuoka1990} & D (ANN3) & C & C & F & L & \unavailable
    \\\hdashline
    \newSkeMethodIndex & \citesp{Hayashi90} & D (ANN) & C & B & F & L & \unavailable
    \\\hdashline
    \newSkeMethodIndex & \citesp{TowellS91} & D (ANN) & C & D & MN & L & \unavailable
    \\\hdashline
    \newSkeMethodIndex & \citesp{Berenji91} & D (ANN) & C & C & F & L & \unavailable
    \\\hdashline
    \newSkeMethodIndex & \citesp{BrunkP91} & P & C & C+D & P & L & \unavailable
    \\\hdashline
    \newSkeMethodIndex & \citesp{murphy1991id2} & P & C & D & MN & DT & \unavailable
    \\\hdashline
    \newSkeMethodIndex & \citesp{HorikawaFU92} & D (ANN) & C & C & F & L & \unavailable
    \\\hdashline
    \newSkeMethodIndex & \citesp{TrespHA92} & D (ANN) & R & C & P & L & \unavailable
    \\\hdashline
    \newSkeMethodIndex & \citesp{towell1993extracting} & D (ANN) & C & D & P & L & \unavailable
    \\\hdashline
    \newSkeMethodIndex & \citesp{Thrun1993ExtractingPC} & D (ANN) & C & C & P+MN & L & \unavailable
    \\\hdashline
    \newSkeMethodIndex & \citesp{Cohen93} & P & C & C+D & P & L & \unavailable
    \\\hdashline
    \newSkeMethodIndex & \citesp{quinlan1993c4} & P & C & C+D & P & DT & \libavailable\footnote{\url{https://en.wikipedia.org/wiki/C4.5\_algorithm\#Implementations}}
    \\\hdashline
    \newSkeMethodIndex & \citesp{Fu94} & D (ANN) & C & D & P & L & \unavailable
    \\\hdashline
    \newSkeMethodIndex & \citesp{halgamuge1994neural} & D (ANN) & C & C & F & L & \unavailable
    \\\hdashline
    \newSkeMethodIndex & \citesp{MITRA1994285} & D (ANN) & C & C+D & F & L & \unavailable
    \\\hdashline
    \newSkeMethodIndex & \citesp{craven1994using} & P & C & B & P+MN & L & \available$^{\ref{foot:psyke}}$
    \\\hdashline
    \newSkeMethodIndex & \citesp{FurnkranzW94} & P & C & D & P & L & \available\footnote{\url{https://github.com/buoto/irep-rule-induction}}
    \\\hdashline
    \newSkeMethodIndex & \citesp{sestito94automated} & P & C & C+D & P & L & \unavailable
    \\\hdashline
    \newSkeMethodIndex & \citesp{PopTHSD95} & P & C & B & P & L & \unavailable
    \\\hline
    \newSkeMethodIndex & \citesp{Andrews95rulex} & D (ANN) & C & C+D & P & L & \unavailable
    \\\hdashline
    \newSkeMethodIndex & \citeauthor{Matthews95fuzzy} (I) \cite{Matthews95fuzzy} & \citeyear{Matthews95fuzzy} & D (ANN) & C & B & F & L & \unavailable
    \\\hdashline
    \newSkeMethodIndex & \citeauthor{Matthews95fuzzy} (II) \cite{Matthews95fuzzy} & \citeyear{Matthews95fuzzy} & D (ANN) & C & B & F & L & \unavailable
    \\\hdashline
    \newSkeMethodIndex & \citesp{Cohen95} & P & C & C+D & P & L & \libavailable\footnote{\url{https://github.com/imoscovitz/wittgenstein}}
    \\\hdashline
    \newSkeMethodIndex & \citesp{SetionoL96} & D (ANN3) & C & B & P & L & \unavailable
    \\\hdashline
    \newSkeMethodIndex & \citesp{tickle1996dedec} & P & C & B & P & L & \unavailable
    \\\hdashline
    \newSkeMethodIndex & \citesp{YuanZ96} & P & C & D & F & L & \unavailable
    \\\hdashline
    \newSkeMethodIndex & \citesp{craven1996extracting} & P & C & B & P+MN & DT & \available\footnote{\url{https://github.com/abarthakur/trepan\_python}}$^{,\ref{foot:psyke}}$
    \\\hdashline
    \newSkeMethodIndex & \citesp{HongL96} & P & C & C & F & L & \unavailable
    \\\hdashline
    \newSkeMethodIndex & \citesp{Setiono97NeuroLinear} & D (ANN3) & C & C+D & O & L & \unavailable
    \\\hdashline
    \newSkeMethodIndex & \citesp{Setiono97a} & D (ANN3) & C & D & P & L & \unavailable
    \\\hdashline
    \newSkeMethodIndex & \citesp{NauckK97} & D (ANN) & C & D & F & L & \unavailable
    \\\hdashline
    \newSkeMethodIndex & \citesp{SaitoN97} & D (ANN) & R & C & P & L & \unavailable
    \\\hdashline
    \newSkeMethodIndex & \citesp{BenitezCR97} & D (ANN) & C & C & F & L & \unavailable
    \\\hdashline
    \newSkeMethodIndex & \citesp{IshibuchiNM97} & P & C & C & F & L & \unavailable
    \\\hdashline
    \newSkeMethodIndex & \citesp{TahaG99} & D (ANN) & C & C & P & L & \unavailable
    \\\hdashline
    \newSkeMethodIndex & \citesp{TahaG99} & D (ANN) & C & C+D & P & L & \unavailable
    \\\hdashline
    \newSkeMethodIndex & \citesp{Krishnan99combo} & D (ANN) & C & B & P & L & \unavailable
    \\\hdashline
    \newSkeMethodIndex & \citesp{NauckK99} & D (ANN) & R & D & F & L & \available\footnote{\url{http://fuzzy.cs.ovgu.de/nefprox/}}
    \\\hdashline
    \newSkeMethodIndex & \citesp{TahaG99} & P & C & B & P & L & \unavailable
    \\\hdashline
    \newSkeMethodIndex & \citesp{KrishnanSB99} & P & C & C & P & DT & \unavailable
    \\\hdashline
    \newSkeMethodIndex & \citesp{schmitz1999ann} & P & C+R & C+D & P & DT & \libavailable\footnote{\label{foot:ruleex}\url{ https://github.com/fantamat/ruleex}}
    \\\hdashline
    \newSkeMethodIndex & \citesp{HongC99} & P & C & C & F & L & \unavailable
    \\\hline
    \newSkeMethodIndex & \citesp{Setiono00} & D (ANN3) & C & B & MN & L & \unavailable
    \\\hdashline
    \newSkeMethodIndex & \citesp{Tsukimoto00} & D (ANN) & C & C+D & P & L & \unavailable
    \\\hdashline
    \newSkeMethodIndex & \citesp{Kim2000} & D (ANN4) & C & C+D & P & DT & \unavailable
    \\\hdashline
    \newSkeMethodIndex & \citesp{SetionoL00} & D (ANN) & R & C+D & P+MN+O & DT & \unavailable
    \\\hdashline
    \newSkeMethodIndex & \citesp{ZhouCC00} & P & C & C+D & P & L & \unavailable
    \\\hdashline
    \newSkeMethodIndex & \citesp{HongC00a} & P & C & C & F & L & \unavailable
    \\\hdashline
    \newSkeMethodIndex & \citesp{sato2001rule} & D (ANN3) & C & C+D & P & DT & \available\footnote{\url{https://github.com/zju-vipa/awesome-neural-trees}}
    \\\hdashline
    \newSkeMethodIndex & \citesp{parpinelli2001ant} & P & C & C+D & P & L & \libavailable\footnote{\url{https://github.com/febo/myra}}
    \\\hdashline
    \newSkeMethodIndex & \citesp{CastilloGP01} & P & C+R & C+D & F & L & \unavailable
    \\\hdashline
    \newSkeMethodIndex & \citesp{saito2002extracting} & D (ANN) & R & C+D & P & L & \unavailable
    \\\hdashline
    \newSkeMethodIndex & \citesp{setiono2002extraction} & D (ANN3) & R & C+D & P & L & \unavailable
    \\\hdashline
    \newSkeMethodIndex & \citesp{liu2002density} & P & C & C+D & P & L & \unavailable
    \\\hdashline
    \newSkeMethodIndex & \citesp{Boz02} & P & C & C+D & P & DT & \unavailable
    \\\hdashline
    \newSkeMethodIndex & \citesp{Markowska-KaczmarT03} & P & C & C+D & F & L & \unavailable
    \\\hdashline
    \newSkeMethodIndex & \citesp{ZhouJC03} & P & C & C+D & P & L & \unavailable
    \\\hdashline
    \newSkeMethodIndex & \citesp{SetionoT04} & D (ANN3) & R & C+D & P & L & \unavailable
    \\\hdashline
    \newSkeMethodIndex & \citesp{Fu2004} & D (SVM) & C & C+D & P & L & \unavailable
    \\\hdashline
    \newSkeMethodIndex & \citesp{Markowska-KaczmarC04} & P & C & C+D & P & L & \unavailable
    \\\hdashline
    \newSkeMethodIndex & \citesp{RabunalDPPR04} & P & C & C+D & P & L & \unavailable
    \\\hdashline
    \newSkeMethodIndex & \citesp{Chen2004LEARNINGAA} & P & C & C & P & L & \unavailable
    \\\hdashline
    \newSkeMethodIndex & \citesp{LiuAM04} & P & C & C+D & P & L & \available\footnote{\url{https://rdrr.io/github/adriansidor/antminer/src/R/antminer3.R}}
    \\\hdashline
    \newSkeMethodIndex & \citesp{Browne2004} & P & C & C+D & P+MN & DT & \unavailable
    \\\hline
    \newSkeMethodIndex & \citesp{ZhangSJC05} & D (SVM) & C & C & P & L & \unavailable
    \\\hdashline
    \newSkeMethodIndex & \citesp{barakat2005eclectic} & D (SVM) & C & C & P & DT & \unavailable
    \\\hdashline
    \newSkeMethodIndex & \citesp{FungSR05} & D (LC) & C & C & P & L & \unavailable
    \\\hdashline
    \newSkeMethodIndex & \citesp{ChavesVT05} & D (SVM) & C & C & F & L & \unavailable
    \\\hdashline
    \newSkeMethodIndex & \citesp{Torres2005} & P & C & C+D & P+MN & DT & \unavailable
    \\\hdashline
    \newSkeMethodIndex & \citesp{EtchellsL06} & P & C & C+D & P & L & \unavailable
    \\\hdashline
    \newSkeMethodIndex & \citesp{He2006} & P & C & C+D & P & DT & \unavailable
    \\\hdashline
    \newSkeMethodIndex & \citesp{huysmans2006iter} & P & R & C & P & L & \available$^{\ref{foot:psyke}}$
    \\\hdashline
    \newSkeMethodIndex & \citesp{BaderHM07} & D (ANN) & C & B & P & L & \unavailable
    \\\hdashline
    \newSkeMethodIndex & \citesp{SchetininFPCKEBH07} & D (DTE) & C & C+D & P & DT & \unavailable
    \\\hdashline
    \newSkeMethodIndex & \citesp{ChenLW07} & D (SVM) & C & C & P & L & \unavailable
    \\\hdashline
    \newSkeMethodIndex & \citesp{BarakatB07} & D (SVM) & C & C+D & P & L & \unavailable
    \\\hdashline
    \newSkeMethodIndex & \citesp{SaadW07} & P & C & C+D & O & L & \available$^{\ref{foot:ruleex}}$
    \\\hdashline
    \newSkeMethodIndex & \citesp{MartensBHVSB07} & P & C & C+D & P & L & \unavailable
    \\\hdashline
    \newSkeMethodIndex & \citesp{NunezAC08} & D (SVM) & C & C & P+O & L & \unavailable
    \\\hdashline
    \newSkeMethodIndex & \citesp{SetionoBM08} & P & C & C+D & P+O & L & \unavailable
    \\\hdashline
    \newSkeMethodIndex & \citesp{OdajimaHTS08} & P & C & D & P & L & \unavailable
    \\\hdashline
    \newSkeMethodIndex & \citesp{grex-icdm2008} & P & C+R & C+D & P+F & DT & \unavailable
    \\\hdashline
    \newSkeMethodIndex & \citesp{Bader09} & D (ANN) & C & B & P & L & \unavailable
    \\\hdashline
    \newSkeMethodIndex & \citesp{MartensBG09} & D (SVM) & C & C+D & any & any & \unavailable
    \\\hline
    \newSkeMethodIndex & \citesp{LehmannBH10} & P & C & B & P & L & \unavailable
    \\\hdashline
    \newSkeMethodIndex & \citesp{AugastaK12} & P & C & C+D & P & L & \unavailable
    \\\hdashline
    \newSkeMethodIndex & \citesp{sethi2012kdruleex} & P & C & C+D & P & TA & \unavailable
    \\\hline
    \newSkeMethodIndex & \citesp{ZilkeMJ16} & D (ANN) & C & C+D & P & DT & \available$^{\ref{foot:ruleex}}$
    \\\hdashline
    \newSkeMethodIndex & \citesp{ChanC17} & D (ANN) & R & C & P & L & \unavailable
    \\\hdashline
    \newSkeMethodIndex & \citesp{BiswasCPRT17} & P & C & C+D & P & L & \unavailable
    \\\hdashline
    \newSkeMethodIndex & \citesp{YedjourB18} & P & C & B & P & L & \unavailable
    \\\hdashline
    \newSkeMethodIndex & \citesp{ChakrabortyBP18} & P & C & C+D & P+O & L & \unavailable
    \\\hdashline
    \newSkeMethodIndex & \citesp{ObregonKJ19} & D (DTE) & C & C+D & P & L & \unavailable
    \\\hline
    \newSkeMethodIndex & \citesp{CHAN2020329} & D (ANN) & R & C & P & L & \unavailable
    \\\hdashline
    \newSkeMethodIndex & \citesp{WangWWYWJ20} & D (DTE) & C & C & P & L & \unavailable
    \\\hdashline
    \newSkeMethodIndex & \citesp{ChenBGLL020} & D (ANN) & C & I & P & L & \available\footnote{\url{https://github.com/SeekingDream/FSE20_DENAS}}
    \\\hdashline
    \newSkeMethodIndex & \citesp{MahdavifarG20} & D (ANN) & C & B & P & L & \unavailable
    \\\hdashline
    \newSkeMethodIndex & \citesp{VasilevMN20} & D (ANN) & C & C+D & P & DT & \unavailable
    \\\hdashline
    \newSkeMethodIndex & \citesp{Odense2020Layerwise} & D (ANN) & C & I & MN & L & \unavailable
    \\\hdashline
    \newSkeMethodIndex & \citesp{JiaLLZL20} & D (ANN) & C & I & P & DT & \unavailable
    \\\hdashline
    \newSkeMethodIndex & \citesp{LiZMZQ20} & D (ANN) & C & C & F & L & \unavailable
    \\\hdashline
    \newSkeMethodIndex & \citesp{Hayashi2020One} & D (ANN) & C & C+D & P & L & \unavailable
    \\\hdashline
    \newSkeMethodIndex & \citesp{ChakrabortyBP20} & D (ANN) & C & C+D & P & L & \unavailable
    \\\hdashline
    \newSkeMethodIndex & \citesp{gridex-extraamas2021} & P & R & C & P & L & \available$^{\ref{foot:psyke}}$
    \\\hdashline
    \newSkeMethodIndex & \citesp{YuNn2021} & D (ANN) & C & C & P & L & \unavailable
    \\\hdashline
    \newSkeMethodIndex & \citesp{YanCZPYHZY21} & D (ANN) & C & C & P & DT & \unavailable
    \\\hdashline
    \newSkeMethodIndex & \citesp{DattachaudhuriB21} & P & C & C+D & P & L & \unavailable
    \\\hdashline
    \newSkeMethodIndex & \citesp{DongYY21} & D (DTE) & C & C+D & P & L & \unavailable
    \\\hdashline
    \newSkeMethodIndex & \citesp{shams2021rem} & D (ANN) & C & C & P & L & \libavailable\footnote{\label{foot:remix}\url{https://github.com/mateoespinosa/remix}}
    \\\hdashline
    \newSkeMethodIndex & \citesp{Yedjour2021Genetic} & P & C & C+D & P & L & \unavailable
    \\\hdashline
    \newSkeMethodIndex & \citesp{marshakov2021rule} & D (ANN) & C & C & F & L & \unavailable
    \\\hdashline
    \newSkeMethodIndex & \citesp{cheng2021} & D (GNN) & C & G & KG & T & \available\footnote{\url{https://github.com/BUPT-GAMMA/CPF}}
    \\\hdashline
    \newSkeMethodIndex & \citesp{recon2021} & D (GNN) & C & T & KG & T & \available\footnote{\url{https://github.com/ansonb/RECON}}
    \\\hdashline
    \newSkeMethodIndex & \citesp{horta2021} & D (ANN) & C & I & KG & T & \unavailable
    \\\hdashline
    \newSkeMethodIndex & \citesp{Bologna21} & D (DTE) & C & C & P & L & \unavailable
    \\\hdashline
    \newSkeMethodIndex & \citesp{EspinosaZarlenga21ECLAIRE} & D (ANN) & C & C & P & L & \available$^{\ref{foot:remix}}$
    \\\hdashline
    \newSkeMethodIndex & \citesp{gridrex-kr2022} & P & R & C & P & L & \available$^{\ref{foot:psyke}}$
    \\\hdashline
    \newSkeMethodIndex & \citesp{guarantees2022} & P & R & C & P & DT & \unavailable
    \\\hdashline
    \newSkeMethodIndex & \citesp{lens2022} & D (ANN) & C & I & P & L & \libavailable\footnote{\url{https://github.com/pietrobarbiero/pytorch_explain}}
    \\\hdashline
    \newSkeMethodIndex & \citesp{ferreira2022} & D (ANN) & C & I & P & L & \unavailable
    \\\hdashline
    \newSkeMethodIndex & \citesp{crin2022} & D (ANN4) & C & C+D & P & L & \unavailable
    \\\hdashline
    \newSkeMethodIndex & \citesp{Barbado2022} & D (SVM) & C & C+D & P & L & \libavailable\footnote{\url{https://github.com/AlbertoBarbado/rule_extraction_xai}}
    \\\hdashline
    \newSkeMethodIndex & \citesp{efnn-nulluni2022} & D (FNN) & C & C+D & F & L & \unavailable
    \\\hdashline
    \newSkeMethodIndex & \citesp{tsk-icfnn2022} & D (FNN) & R & C & F & L & \unavailable
    \\\hdashline
    \newSkeMethodIndex & \citesp{calcs-cl2022} & D (CNN) & C & I & P & L & \unavailable
    \\\hdashline
	\newSkeMethodIndex & \citesp{creepy2023beware} & P & C+R & C & P & DT & \available$^{\ref{foot:psyke}}$
	\\\hdashline
    \newSkeMethodIndex & \citesp{rulecosi2023} & D (DTE) & C & C+D & P & L & \available\footnote{\url{https://github.com/jobregon1212/rulecosi}}
    \\\hdashline
    \newSkeMethodIndex & \citesp{len2023} & P & C & C+D & P & L & \libavailable\footnote{\url{https://github.com/pietrobarbiero/logic_explained_networks}}
\end{longtable}


\subsection{Summary about SKI}
\label{app:summary-ski}

\Cref{tab:ski-taxonomy} summarises our analysis regarding the \skicount{} surveyed SKI methods.
Notably, the table enumerates SKI methods in chronological order (w.r.t.\ publication year), grouping them by five-year periods.
Furthermore, coherently w.r.t.\ the sections above, the table reports the strategy followed by each SKI method (see \Cref{subsubsec:RQ9}), as well as the type of knowledge it can inject (see \Cref{subsubsec:RQ7}), the type of neural network it supports (see \Cref{subsubsec:RQ6}), and the overall purpose it supports injection for (see \Cref{subsubsec:RQ8}).
The availability/lack of a public software implementation of each SKI algorithm is also reported.


\newcommand{\skitabhead}{
	\textbf{\#} & \textbf{Method} & \textbf{Year} & \makecell{\textbf{Strategy}\\(\ref{item:rq9})} & \makecell{\textbf{Input}\\(\ref{item:rq7})} & \makecell{\textbf{Predictor}\\(\ref{item:rq6})} & \makecell{\textbf{Purpose}\\(\ref{item:rq8})} & \makecell{\textbf{Tech.}\\(\ref{item:rq10})} \\
	\hline\hline
}

\newcounter{SkiMethod}
\setcounter{SkiMethod}{1}
\newcommand{\newSkiMethodIndex}{\theSkiMethod\stepcounter{SkiMethod}}

\begin{longtable}{c|p{4cm}|c|c|c|c|c|c}
    \caption{
        Summary of knowledge-injection algorithms.
        %
        %
        Values from the columns ``Strategy'', ``Input'', ``Predictor'', ``Purpose'', and ``Technology'' refer the corresponding figures form Sec. 4.2.
        %
    }
    \label{tab:ski-taxonomy}\\        
    \skitabhead
    \endfirsthead
    \caption[]{Summary of knowledge-injection algorithms (continued).}\\
    \skitabhead
    \endhead
    \endlastfoot
    \newSkiMethodIndex & \citesp{BallardAaai1986} & S & FOL & BM & M & \unavailable
    \\\hline
    \newSkiMethodIndex & \citesp{TowellAaai1990} & S & P & FF & E & \libavailable\footnote{\label{foot:psyki}\url{https://github.com/psykei/psyki-python}}
    \\\hdashline
    \newSkiMethodIndex & \citesp{PinkusNips1991} & S & FOL & BM & M & \unavailable
    \\\hdashline
    \newSkiMethodIndex & \citesp{TrespNips1992} & L+S & P & FF & E+M & \unavailable
    \\\hdashline
    \newSkiMethodIndex & \citesp{GilesIcnn1993} & S & E & RNN & E+M & \unavailable
    \\\hline
    \newSkiMethodIndex & \citesp{TanTnn1997} & S & P & FF & E & \unavailable
    \\\hdashline
    \newSkiMethodIndex & \citesp{GarcezAppint1999} & S & P & FF & M & \libavailable\footnote{\url{https://sourceforge.net/projects/cil2p/}}
    \\\hline
    \newSkiMethodIndex & \citesp{BasilioIlp2001} & L+S & FOL & FF & M & \unavailable
    \\\hdashline
    \newSkiMethodIndex & \citesp{GarcezAaai2004} & S & FOL & FF & M & \unavailable
    \\\hdashline
    \newSkiMethodIndex & \citesp{BaderFlairs2005} & S & FOL & FF & M & \unavailable
    \\\hline
    \newSkiMethodIndex & \citesp{ChangAcl2007} & E & E & MN & E & \unavailable
    \\\hdashline
    \newSkiMethodIndex & \citesp{BaderNesy2008} & L & E & FF & E & \unavailable
    \\\hdashline
    \newSkiMethodIndex & \citesp{MintzAcl2009} & E & KG & FF & M & \unavailable
    \\\hline
    \newSkiMethodIndex & \citesp{NickelIcml2011} & E & KG & FF & M & \libavailable\footnote{\url{https://github.com/mnick/rescal.py}}
    \\\hdashline
    \newSkiMethodIndex & \citesp{BordesWCB11} & E & KG & FF & M & \available\footnote{\url{https://github.com/glorotxa/SME}}
    \\\hdashline 
    \newSkiMethodIndex & \citesp{KimmigNips2012} & S & FOL & MN & M & \libavailable\footnote{\url{https://github.com/linqs/psl}}
    \\\hdashline
    \newSkiMethodIndex & \citesp{BordesGWB12} & E & KG & FF & M & \unavailable
    \\\hdashline
    \newSkiMethodIndex & \citesp{PinkasBica2013} & S & FOL & BM & M & \unavailable
    \\\hdashline
    \newSkiMethodIndex & \citesp{BordesNips2013} & E+L & KG & FF & M & \available\footnote{\url{https://github.com/Lapis-Hong/TransE-Knowledge-Graph-Embedding}}
    \\\hdashline
    \newSkiMethodIndex & \citesp{SocherCMN13} & E+S & KG & FF & M & \available\footnote{\url{https://github.com/dddoss/tensorflow-socher-ntn}}
    \\\hdashline
    \newSkiMethodIndex & \citesp{FrancaMl2014} & S & P & RNN & M & \available\footnote{\url{https://github.com/vakker/CILP}}
    \\\hdashline
    \newSkiMethodIndex & \citesp{WangAaai2014} &  E+L & KG & FF & M & \available\footnote{\label{foot:KB2E}\url{https://github.com/thunlp/KB2E}}
    \\\hdashline
    \newSkiMethodIndex & \citesp{GarciaduranEcml2014} & E+L & KG & FF & M & \unavailable
    \\\hdashline
    \newSkiMethodIndex & \citesp{BianEcml2014} & E+L & E & FF & E & \unavailable
    \\\hdashline
    \newSkiMethodIndex & \citesp{ChangEmnlp2014} &  E & KG & FF & M & \unavailable
    \\\hdashline
    \newSkiMethodIndex & \citesp{BordesGWB14} & E & KG & FF & M & \available\footnote{\url{https://github.com/usherwang02/SemanticMatchingEnergy-Theano}}
    \\\hdashline
    \newSkiMethodIndex & \citesp{0001GHHLMSSZ14} & E & KG & FF & M & \unavailable
    \\\hdashline
    \newSkiMethodIndex & \citesp{paclic/FanZCZ14} & E+L & KG & FF & M & \unavailable
    \\\hline
    \newSkiMethodIndex & \citesp{WangIjcai2015} & E & KG & FF & M & \unavailable
    \\\hdashline
    \newSkiMethodIndex & \citesp{WeiCikm2015} & E & KG & MN & M & \available\footnote{\url{https://github.com/ZhuoyuWei/fpMLN}}
    \\\hdashline
    \newSkiMethodIndex & \citesp{RocktaschelNaactl2015} & E+L & KG & FF & M & \available\footnote{\url{https://github.com/uclnlp/low-rank-logic}}
    \\\hdashline
    \newSkiMethodIndex & \citesp{LinAaai2015} &  E+L & KG & FF & M & \available$^{\ref{foot:KB2E}}$
    \\\hdashline
    \newSkiMethodIndex & \citesp{YangIclr2015} &  E+L & KG & FF & M & \unavailable
    \\\hdashline
    \newSkiMethodIndex & \citesp{CheKdd2015} &  L & KG & FF & E & \unavailable
    \\\hdashline
    \newSkiMethodIndex & \citesp{JiHXL015} & E+L & KG & FF & M & \unavailable
    \\\hdashline
    \newSkiMethodIndex & \citesp{FengHWZHZ16} & E+L & KG & FF & M & \unavailable
    \\\hdashline
    \newSkiMethodIndex & \citesp{0005HHZ15a} & E+L & KG & FF & M & \unavailable
    \\\hdashline
    \newSkiMethodIndex & \citesp{HeLJ015} & E+L & KG & FF & M & \unavailable
    \\\hdashline
    \newSkiMethodIndex & \citesp{TranTnnls2016} & S & P & DBN & E & \unavailable
    \\\hdashline
    \newSkiMethodIndex & \citesp{HuAcl2016} & S & P & CNN & E & \unavailable
    \\\hdashline
    \newSkiMethodIndex & \citesp{GuoEmnlp16} & E+L & KG & FF & M & \unavailable
    \\\hdashline
    \newSkiMethodIndex & \citesp{NickelAaai2016} &  E+L & KG & FF & M & \available\footnote{\url{https://github.com/mnick/holographic-embeddings}}
    \\\hdashline
    \newSkiMethodIndex & \citesp{TrouillonIcml2016} & E+L & KG & FF & M & \available \footnote{\url{https://github.com/thunlp/openke}}
    \\\hdashline
    \newSkiMethodIndex & \citesp{DemeesterEmnlp2016} &  L & KG & FF & M & \unavailable
    \\\hdashline
    \newSkiMethodIndex & \citesp{HuEmnlp2016} & S & P & FF & E & \unavailable
    \\\hdashline
    \newSkiMethodIndex & \citesp{MrksicNaacl2016} &  L & KG & FF & E & \available\footnote{\url{https://github.com/nmrksic/counter-fitting}}
    \\\hdashline
    \newSkiMethodIndex & \citesp{Liu0LWH16} & E & KG & FF & M & \unavailable
    \\\hdashline
    \newSkiMethodIndex & \citesp{JiLH016} & E+L & KG & FF & M & \unavailable
    \\\hdashline
    \newSkiMethodIndex & \citesp{acl/0005HZ16} & E+L & KG & FF & M & \unavailable
    \\\hdashline
    \newSkiMethodIndex & \citesp{ijcai/0005HZ16} & E+L & KG & FF & M & \unavailable
    \\\hdashline
    \newSkiMethodIndex & \citesp{KipfIclr2017} &  E+L & KG & GNN & M & \libavailable\footnote{\url{https://github.com/tkipf/pygcn}}
    \\\hdashline
    \newSkiMethodIndex & \citesp{RocktaschelNips2017} & L+S & D & FF & M & \available\footnote{\url{https://github.com/uclnlp/ntp}}
    \\\hdashline
    \newSkiMethodIndex & \citesp{LiuIcml2017} &  E+L & KG & FF & M & \available\footnote{\url{https://github.com/quark0/ANALOGY}}
    \\\hdashline
    \newSkiMethodIndex & \citesp{StewartAaai2017} & L & E & CNN & E & \unavailable
    \\\hdashline
    \newSkiMethodIndex & \citesp{AllamanisIcml2017} & L & P & RNN & E & \available\footnote{\url{https://github.com/mast-group/eqnet}}
    \\\hdashline
    \newSkiMethodIndex & \citesp{DiligentiAi2017} & L & FOL & KM & M & \available\footnote{\url{https://sites.google.com/site/semanticbasedregularization/home/software}}
    \\\hdashline
    \newSkiMethodIndex & \citesp{DiligentiIcmla2017} & L & P & CNN & M & \unavailable
    \\\hdashline
    \newSkiMethodIndex & \citesp{MarinoCvpr2017} &  E & KG & GNN & E & \unavailable
    \\\hdashline
    \newSkiMethodIndex & \citesp{ChangIclr2017} & E & E & FF & E & \available\footnote{\url{https://github.com/mbchang/dynamics}}
    \\\hdashline
    \newSkiMethodIndex & \citesp{ChoiKdd2017} & E & KG & FF & E & \available\footnote{\url{https://github.com/mp2893/gram}}
    \\\hdashline
    \newSkiMethodIndex & \citesp{FangIjcai2017} & L & KG & CNN & E & \unavailable
    \\\hdashline
    \newSkiMethodIndex & \citesp{XuIcml2018} &  L & P & CNN & E & \available\footnote{\url{https://github.com/UCLA-StarAI/Semantic-Loss/}}
    \\\hdashline
    \newSkiMethodIndex & \citesp{EvansJair2018} & L+S & D & FF & M & \available\footnote{\url{https://github.com/crunchiness/lernd}}
    \\\hdashline
    \newSkiMethodIndex & \citesp{SourekJair2018} & S & D & FF & M & \available\footnote{\url{https://github.com/GustikS/GNNwLRNNs}}
    \\\hdashline
    \newSkiMethodIndex & \citesp{VelickovicIclr2018} &  E+L & KG & GNN & M & \available\footnote{\url{https://github.com/PetarV-/GAT}}
    \\\hdashline
    \newSkiMethodIndex & \citesp{MaBibm2018} &  L & KG & AE & E & \unavailable
    \\\hdashline
    \newSkiMethodIndex & \citesp{ZhouIjcai2018} & E & KG & GNN & E & \available\footnote{\url{https://github.com/tuxchow/ccm}}
    \\\hdashline
    \newSkiMethodIndex & \citesp{LiangNips2018} &  S & KG & FF & E & \available\footnote{\url{https://github.com/julianschoep/SGRLayer}}
    \\\hdashline
    \newSkiMethodIndex & \citesp{GlavasAcl2018} &  E+L & KG & FF & E & \available\footnote{\url{https://github.com/codogogo/explirefit}}
    \\\hdashline
    \newSkiMethodIndex & \citesp{MarraCorr2019} &L & P & FF & E & \available\footnote{\url{https://github.com/GiuseppeMarra/lyrics}}
    \\\hdashline
    \newSkiMethodIndex & \citesp{GoodwinCorr2019} &  L & KG & FF & E & \unavailable
    \\\hdashline
    \newSkiMethodIndex & \citesp{SunIclr2019} &  E+L & KG & FF & M & \unavailable
    \\\hdashline
    \newSkiMethodIndex & \citesp{ZhangAcl2019} &  L & KG & TR & E & \unavailable
    \\\hdashline
    \newSkiMethodIndex & \citesp{PetersEmnlp2019} &  E+L & KG & TR & E & \libavailable\footnote{\url{https://github.com/allenai/kb}}
    \\\hdashline
    \newSkiMethodIndex & \citesp{DanielePricai2019} &  S & FOL & DFF & E & \libavailable\footnote{\url{https://github.com/DanieleAlessandro/KENN}}
    \\\hdashline
    \newSkiMethodIndex & \citesp{FischerIcml2019} &  L & D & DFF & E & \available\footnote{\url{https://github.com/eth-sri/dl2}}
    \\\hdashline
    \newSkiMethodIndex & \citesp{DongIclr2019} &  S+L & H & FF & M & \available\footnote{\url{https://github.com/google/neural-logic-machines}}
    \\\hline
    \newSkiMethodIndex & \citesp{BadreddineCorr2020} & S & FOL & FF & E+M & \libavailable\footnote{\url{https://github.com/logictensornetworks/logictensornetworks}}
    \\\hdashline
    \newSkiMethodIndex & \citesp{ZhangAaai2020} &  E+L & KG & FF & M & \available\footnote{\url{https://github.com/MIRALab-USTC/KGE-HAKE}}
    \\\hdashline
    \newSkiMethodIndex & \citesp{JiangArtmed2020} &  S+L & FOL & RNN & E & \unavailable
    \\\hdashline
    \newSkiMethodIndex & \citesp{RenNips2020} &  S+L & KG & DFF & M & \available\footnote{\url{https://github.com/snap-stanford/KGReasoning}}
    \\\hdashline
    \newSkiMethodIndex & \citesp{GuoCikm2020} &  L+E & KG & FF & M & \available\footnote{\url{https://github.com/StudyGroup-lab/SLRE}}
    \\\hdashline
    \newSkiMethodIndex & \citesp{RiegelCorr2020} &  S+L & FOL & FF & M & \libavailable\footnote{\url{https://github.com/IBM/LNN}}
    \\\hdashline
    \newSkiMethodIndex & \citesp{YuNn2021} & S & P & DAE & E & \unavailable
    \\\hdashline
    \newSkiMethodIndex & \citesp{ManhaeveAi2021} & S & H & FF & E+M & \libavailable\footnote{\url{https://github.com/ML-KULeuven/deepproblog}}
    \\\hdashline
    \newSkiMethodIndex & \citesp{DashMl2021} & E & P & GNN & E & \available\footnote{\url{https://github.com/tirtharajdash/VEGNN}}
    \\\hdashline
    \newSkiMethodIndex & \citesp{GiunchigliaJair2021} & S+L & P & CNN & E & \available\footnote{\url{https://github.com/EGiunchiglia/C-HMCNN/}}
    \\\hdashline
    \newSkiMethodIndex & \citesp{BosselutAaai2021} & S & KG & TR & M & \unavailable
    \\\hdashline
    \newSkiMethodIndex & \citesp{PengCorr2021} & E & E & TR & E & \unavailable
    \\\hdashline
    \newSkiMethodIndex & \citesp{WestNaacl2022} & L E & KG & TR & E & \unavailable
    \\\hdashline
    \newSkiMethodIndex & \citesp{MarinoCvpr2021} & S+L & KG & TR & E & \libavailable\footnote{\url{https://github.com/facebookresearch/mmf}}
    \\\hdashline
    \newSkiMethodIndex & \citesp{XieIcra2021} & S+L & M & RNN & E & \unavailable
    \\\hdashline
    \newSkiMethodIndex & \citesp{ChengEmclp2021} & L+E & H & FF & M & \unavailable
    \\\hdashline
    \newSkiMethodIndex & \citesp{LiTkde2023} & S+L & D & GNN & M & \unavailable
    \\\hdashline
    \newSkiMethodIndex & \citesp{dAmatoEswc2021} & L+E & KG & FF & M & \available\footnote{\url{https://github.com/Keehl-Mihael/TransROWL-HRS}}
    \\\hdashline
    \newSkiMethodIndex & \citesp{DashMl2022} & S & P & GNN & E & \available\footnote{\url{https://github.com/tirtharajdash/BotGNN}}
    \\\hdashline
    \newSkiMethodIndex & \citesp{RodriguezIf2022} & L & KG & CNN & E & \available\footnote{\url{https://github.com/JulesSanchez/X-NeSyL}}
    \\\hdashline
    \newSkiMethodIndex & \citesp{YuCvpr2022} & S+L & FOL & CNN & E & \unavailable
    \\\hdashline
    \newSkiMethodIndex & \citesp{WeiIjcnlp2022} & S+L & P & GNN & M & \available\footnote{\url{http://github.com/jinnanli/CogKG}}
    \\\hdashline
    \newSkiMethodIndex & \citesp{SmirnovaTkde2022} & L & P & FF & E & \available\footnote{\url{https://github.com/eXascaleInfolab/Nessy_RE}}
    \\\hdashline
    \newSkiMethodIndex & \citesp{kins-cilc2022} & S & D & FF & E & \available$^{\ref{foot:psyki}}$
    \\\hdashline
    \newSkiMethodIndex & \citesp{SpilloRecsys2022} & E & FOL & DFF & E & \available\footnote{\url{https://github.com/giuspillo/RepoNeSyRecSys2022}}
    \\\hdashline
    \newSkiMethodIndex & \citesp{TangCorr2022} & L+E & FOL & RNN & M & \available\footnote{\url{https://github.com/XiaojuanTang/RulE}}
    \\\hdashline
    \newSkiMethodIndex & \citesp{ZhuIcml2022} & L & FOL & GNN & M & \available\footnote{\url{https://github.com/DeepGraphLearning/GNN-QE}}
    \\\hdashline
    \newSkiMethodIndex & \citesp{LiAppint2022} & L+E & KG & GNN & M & \unavailable
    \\\hdashline
    \newSkiMethodIndex & \citesp{SenAaai2022} & S+L & D & FF & M & \unavailable
    \\\hdashline
    \newSkiMethodIndex & \citesp{kill-woa2022} & S+L & D & DFF & E & \available$^{\ref{foot:psyki}}$
    \\\hdashline
    \newSkiMethodIndex & \citesp{WernerCorr2023} & S & FOL & GNN & E & \available\footnote{\url{https://gitlab.inria.fr/tyrex-public/kegnn}}
    \\\hdashline
    \newSkiMethodIndex & \citesp{GianniniAppint2023} & L & FOL & FF & E & \unavailable
    \\\hdashline
    \newSkiMethodIndex & \citesp{CunningtonMl2023} & S+L & D & FF & E & \available\footnote{\url{https://github.com/DanCunnington/FFNSL}}
    \\\hdashline
    \newSkiMethodIndex & \citesp{PourvaliCorr2023} & S+L & FOL & TR & E & \unavailable
    \\\hdashline
    \newSkiMethodIndex & \citesp{AhmedCorr2023} & L & P & FF & E & \available\footnote{\url{https://github.com/UCLA-StarAI/Semantic-Strengthening}}
    \\\hdashline
    \newSkiMethodIndex & \citesp{MarconatoCorr2023} & L & H & DFF & M+E & \available\footnote{\url{https://github.com/ema-marconato/NeSy-CL}}
    \\\hdashline
    \newSkiMethodIndex & \citesp{LiAaai2023} & S+L & KG & TR & E & \available\footnote{\url{https://github.com/senticnet/SKIER}}
    \\\hdashline
    \newSkiMethodIndex & \citesp{LinInffus2023} & S+L & H & TR & E & \unavailable
    \\\hdashline
    \newSkiMethodIndex & \citesp{BaiInfsci2023} & L & M & GNN & M & \unavailable
    \\\hdashline
    \newSkiMethodIndex & \citesp{NguyenEacl2023} & L+E & FOL & FF & M & \available\footnote{\url{https://github.com/nlp-tlp/cyle}}

\end{longtable}

\bibliographystyle{ACM-Reference-Format}
\bibliography{acmcs-skeskislr}

\end{document}